\documentclass[runningheads]{llncs}
\usepackage[T1]{fontenc}
\usepackage{graphicx}
\usepackage[resetlabels]{multibib}
\usepackage{subfiles}
\usepackage{wrapfig}
\usepackage{csquotes}
\usepackage{gensymb}
\usepackage{pifont}
\usepackage{textcomp}
\usepackage[table]{xcolor}
\usepackage{todonotes}
\usepackage{blindtext}
\usepackage{soul}
\usepackage{graphicx}
\usepackage{mathtools}
\usepackage{multirow}
\usepackage{booktabs}
\usepackage{array}
\usepackage{tabularx}
\usepackage[hang,flushmargin,symbol]{footmisc}
\usepackage{amsmath} 
\usepackage{amssymb} 
\usepackage{amsfonts}
\usepackage{bm} 
\usepackage{siunitx}
\usepackage{cancel}
\usepackage{microtype}
\usepackage{xcolor}
\usepackage[breaklinks,colorlinks]{hyperref}
\usepackage{caption}
\usepackage{subcaption}
\usepackage{graphicx}
\usepackage{cite}
\usepackage[export]{adjustbox}
\usepackage{cuted}
\usepackage{enumitem} 
\usepackage{pgfplots}
\pgfplotsset{width=\linewidth,compat=1.9}
\definecolor{color_red}{RGB}{228,26,28}
\definecolor{color_blue}{RGB}{55,126,184}
\definecolor{color_green}{RGB}{77,175,74}
\definecolor{color_purple}{RGB}{152,78,163}
\definecolor{color_orange}{RGB}{255,127,0}
\definecolor{color_brown}{RGB}{166,86,40}
\definecolor{color_pink}{RGB}{247,129,191}
\newcites{S}{References}
    
\newcommand*{\ARXIV}{}%
\usepackage[capitalize]{cleveref}
\crefname{section}{Sec.}{Secs.}
\Crefname{section}{Section}{Sections}
\Crefname{table}{Table}{Tables}
\crefname{table}{Tab.}{Tabs.}
\definecolor{mygreen}{HTML}{00A64F}
\definecolor{myred}{HTML}{ED1B23}

\newcommand{\secref}[1]{Sec.~\ref{#1}}
\renewcommand{\eqref}[1]{Eq.~(\ref{#1})}
\newcommand{\figref}[1]{Fig.~\ref{#1}}
\newcommand{\tabref}[1]{Tab.~\ref{#1}}

\newcommand{\net}{TOPICS+}
\begin{document}
\title{Dynamic Robot-Assisted Surgery with Hierarchical Class-Incremental Semantic Segmentation}
%
\titlerunning{Hierarchical Class-Incremental Semantic Segmentation}
\ifdefined\ARXIV
\author{Julia Hindel\inst{1} \and
Ema Mekic\inst{1,2} \and
Enamundram Naga Karthik\inst{3,4} \and
Rohit Mohan\inst{1} \and
Daniele Cattaneo\inst{1}\and
Maria Kalweit\inst{1,5} \and
Abhinav Valada\inst{1,2}}

\else
\author{Julia Hindel\inst{1}\orcidID{0000-0002-7228-5717} \and
Ema Mekic\inst{1,2}\orcidID{0009-0008-7842-0749} \and
Enamundram Naga Karthik\inst{3,4}\orcidID{0000-0003-2940-5514} \and
Rohit Mohan\inst{1}\orcidID{0000-0002-5067-4279} \and
Daniele Cattaneo\inst{1}\orcidID{0000-0001-6662-5810} \and
Maria Kalweit\inst{1,5}\orcidID{0000-0003-4581-8810} \and
Abhinav Valada\inst{1,2}\orcidID{0000-0003-4710-3114}}
\fi
\authorrunning{Hindel et al.}
\institute{Department of Computer Science, University of Freiburg, Germany \and Zuse School ELIZA\and NeuroPoly Lab, Institute of Biomedical Engineering, Polytechnique Montréal, Montréal, QC, Canada \and Mila - Québec AI Institute, Montréal, QC, Canada\and Collaborative Research Institute Intelligent Oncology (CRIION)\\
\email{\{hindel, mohan, cattaneo, kalweitm, valada\}@cs.uni-freiburg.de}}
%
\maketitle              
\begin{abstract}
Robot-assisted surgeries rely on accurate and real-time scene understanding to safely guide surgical instruments. However, segmentation models trained on static datasets face key limitations when deployed in these dynamic and evolving surgical environments. Class-incremental semantic segmentation (CISS) allows models to continually adapt to new classes while avoiding catastrophic forgetting of prior knowledge, without training on previous data. 
In this work, we build upon the recently introduced Taxonomy-Oriented Poincaré-regularized
Incremental Class Segmentation (TOPICS) approach and propose an enhanced variant, termed \net, specifically tailored for robust segmentation of surgical scenes.
Concretely, we incorporate the Dice loss into the hierarchical loss formulation to handle strong class imbalances, introduce hierarchical pseudo-labeling, and design tailored label taxonomies for robotic surgery environments. 
We also propose six novel CISS benchmarks designed for robotic surgery environments including multiple incremental steps and several semantic categories to emulate realistic class-incremental settings in surgical environments. In addition, we introduce a refined set of labels with more than $144$ classes on the Syn-Mediverse synthetic dataset, hosted online as an evaluation benchmark. 
We make the code and trained models publicly available at \url{http://topics.cs.uni-freiburg.de}.

\end{abstract}

\section{Introduction}\label{sec:introduction}

In 2024, robotic-assisted surgery accounted for 3\% of surgeries worldwide and continues to grow rapidly, expanding across specialties such as urology, gynecology, orthopedics, dentistry, and neurology. It entails benefits including reduced blood loss (50.5\%), transfusion rates (27.2\%), hospital stays (69.5\%), and 30-day complications (63.7\%)~\cite{biswas23,picozzi24}. However, the complexity and variability of surgical environments call for intelligent scene understanding of surgical instruments and human tissues to enable precise and controlled execution of intricate tasks~\cite{mohan23syn, Xu24}.

Further, hospitals and diagnostic tools need to quickly adapt to new annotation protocols, equipment, and/or patient populations. Conventional deep learning methods 
exhibit catastrophic forgetting~\cite{mccloskey1989catastrophic, vodisch2023covio,valada2016convoluted}, a phenomenon where models forget previously learned knowledge when learning new tasks. In contrast, continual learning (CL) presents a principled approach to \textit{continuously} update models with new incoming data without forgetting prior knowledge, thereby supporting long-term diagnosis required in dynamic clinical settings. Commonly, replay-based CL methods rehearse a subset of prior samples during online training to mitigate forgetting~\cite{karthik2022segmentation, ozdemir2019, ozdemir18}. However, unrestrained access to past data cannot be assumed in clinical scenarios due to privacy regulations and storage constraints~\cite{qazi24}, highlighting the need for non-replay based approaches. Furthermore, class-incremental semantic segmentation (CISS) presents the additional challenge of \textit{background shift}~\cite{cermelli2020mib}, where prior and potential future classes are assigned to the background class, forcing the model to retain recognition despite contradictory supervision.
\subsection{Related Work}
CISS has been explored for multiple modalities in the medical domain. While the earlier method LwF~\cite{li16lwf} was extended with contrastive learning for 3D tract segmentation in neuroimages~\cite{Xu25}, parallel convolutions~\cite{zhang2022representation} were applied to learn new classes under domain shift in brain tumor segmentation~\cite{liu23}. For organ segmentation in 3D CT scans, \cite{liu22} introduce a feature memory buffer to incrementally distinguish five organs. Further work scales 3D organ and body part segmentation to $103$ classes, using separate decoder architectures per organ to mitigate forgetting~\cite{ji23,zhang24miccai}.

For surgical scenes, prior work highlights the benefits of distillation-based approaches~\cite{bai23} for continual Visual-Question Localized-Answering using a two-step incremental setup based on Endovis17~\cite{endovis17} and Endovis18~\cite{endovis18} datasets. GAN-generated backgrounds are also applied for a two-step incremental robotic instrument learning on these datasets~\cite{Xu24}. However, we emphasize that incremental learning scenarios should not be limited to specific categories, and prior work in other domains highlights the exponential increase of forgetting after various steps~\cite{hindel25}. Consequently, we define multiple incremental tasks to evaluate the model's knowledge retention on Endovis18~\cite{endovis18} and introduce benchmark settings on MM-OR~\cite{ozsoy2024mmor} and Syn-Mediverse~\cite{mohan23syn} for large-scale multi-class evaluation.

Further, previous works highlight the suitability of modeling tree-like structures in hyperbolic space~\cite{ganea2018hnn} to mitigate forgetting~\cite{hindel25}. The positive impact of knowledge distillation in hyperbolic spaces has also been studied for continual medical image classification~\cite{roy2023l3dmc}. Building on these works, we tailor the hyperbolic CISS method, TOPICS~\cite{hindel25}, to robotic surgery scenarios for addressing replay-free CISS.

\subsection{Contributions}
This work presents replay-free Class-Incremental Semantic Segmentation (CISS) for continual robot-assisted surgical scene segmentation by building on TOPICS~\cite{hindel25}, a hierarchical, hyperbolic class-incremental learning approach. We integrate the Dice loss~\cite{milletari2016v} into the hierarchical training to address class imbalance in segmenting robotic instruments. Furthermore, we propose hierarchical pseudo-labeling to maintain accuracy across diverse background complexities and define suitable label taxonomies to prevent forgetting. We evaluate our approach across six different CISS settings, including learning disjoint classes and refining existing ones. Our findings show that encoding classes in a hierarchical tree structure helps retain knowledge and generalize to new classes, outperforming non-hierarchical CISS methods. Finally, we extend labels in the Syn-Mediverse dataset~\cite{mohan23syn} for a realistic evaluation, creating over 144 fine categories of surgical environments in hospitals.

In summary, our contributions can be summarized: 
\begin{itemize}[topsep=0pt]
    \item We present \net, a replay-free CISS approach for continual robot-assisted surgical scene segmentation. 
    \item We evaluate CISS methods across six diverse CL settings for robotic surgical environments, demonstrating the superior knowledge retention and generalization of our hierarchical class encoding.
    \item We release fine-grained labels for over 144 categories in surgical environments for the public Syn-Mediverse dataset and host an online evaluation platform for continued benchmarking.
\end{itemize}

\section{Materials and Methods}\label{sec:technical-approach}

\subsection{Datasets}\label{sec:dataset}
We use three datasets, namely, Endovis18~\cite{endovis18}, MM-OR~\cite{ozsoy2024mmor}, and Syn-Mediverse~\cite{mohan23syn} and create novel CISS settings following the protocol described in~\cite{hindel25}. The notation $X$-$Y$($T$ tasks) indicates that the model is initially trained on $X$ base classes at timestep $t=1$ ($C^1$), and subsequently learns $Y$ new classes in each of the remaining $T{-}1$ tasks, resulting in $Y \times (T{-}1)$ novel classes.
We propose different incremental setups where the classes at $t = 2, ..., T$ ($C^{2:T}$) may be disjoint (new classes) or refinement (sub-classes) of base classes $X$ ($t=1$). Every task is defined by its own disjoint label space and training dataset $\mathcal{D}^t$. Note that our CISS settings overlap, i.e., images are composed of pixels from old, current, or future classes. However, only current classes are labeled in $\mathcal{D}^t$ and all remaining pixels are assigned to the background class, which induces a background shift. 

\begin{figure}[t]
  \centering
  \begin{minipage}{0.73\textwidth}
    \centering
    \includegraphics[width=\linewidth]{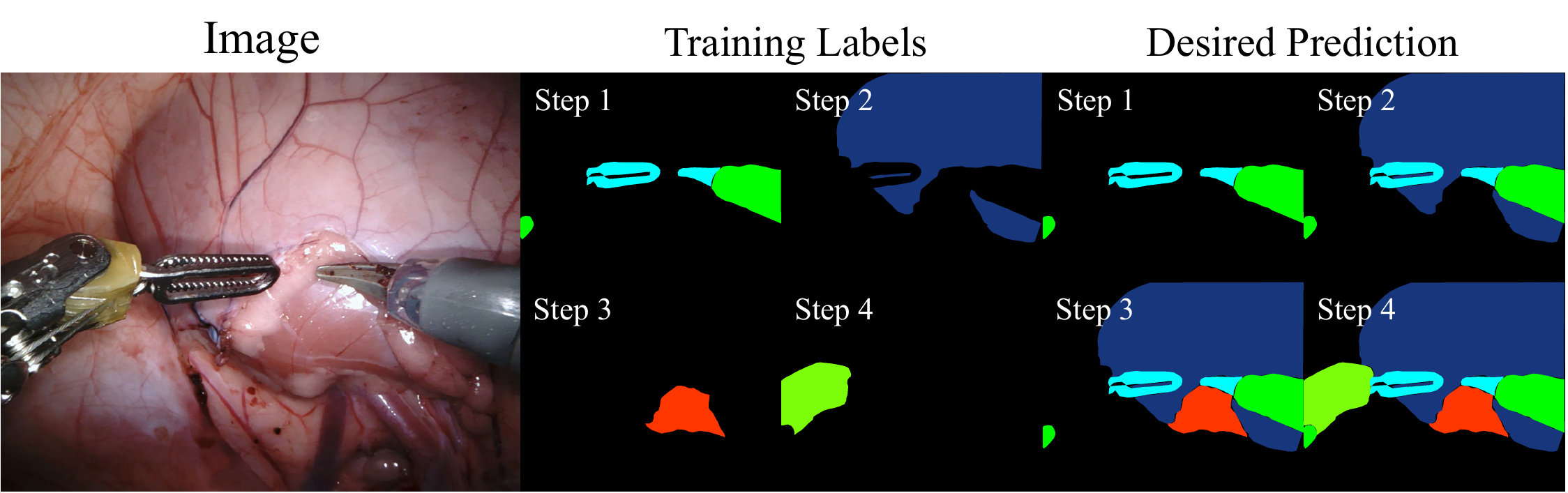}
    \captionof{figure}{CISS on the Endovis18 dataset \cite{endovis18}. Observe the independent training labels at each step and how the model has to learn to segment all classes incrementally. }
    \label{fig:ciss}
  \end{minipage}
  \hfill
  \begin{minipage}{0.24\textwidth}
    \centering
    \includegraphics[width=\linewidth]{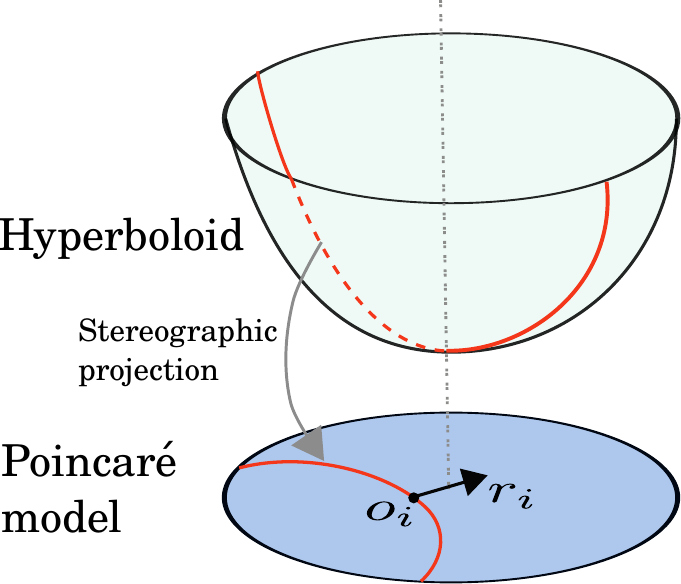}
    \captionof{figure}{Poincaré hyperplane: offset o$_i$, orientation r$_i$~\cite{hindel25}.}
    \label{fig:pc}
  \end{minipage}
\end{figure}

\begin{figure}[t]
    \includegraphics[width=\linewidth]{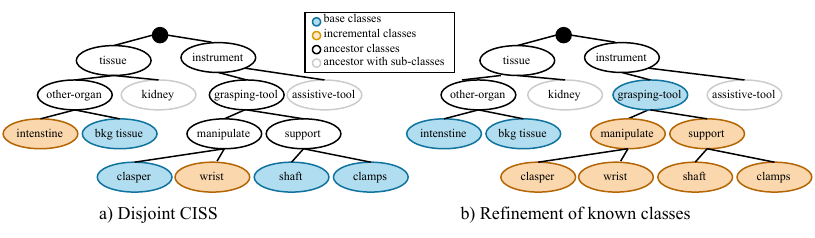}
    \caption{Visualization of CISS settings on the Endovis18 dataset. a) In disjoint CISS, novel classes (orange) originate from the background and are disjoint from base classes (blue). b) Novel classes (orange) are refinements of base classes (blue).}
    \label{fig:tree}
\vspace{-0.5cm}
\end{figure}

{\parskip=3pt\noindent\textbf{Endovis18}: The Endovis18 dataset~\cite{endovis18} categorizes the da Vinci instruments and various anatomical classes from $10$ surgical videos into $12$ semantic categories. We split the dataset into \num{1788} training, \num{447} validation, and \num{997} test images, with the test set representing the union of the four official challenge test sets. 
Following the CISS notation, we design a disjoint $8$-$1$ ($4$ tasks) setup, starting with 8 base classes at $t=1$ and incrementally learning the (i) covered-kidney, (ii) kidney-parenchyma, (iii) instrument-wrist, and (iv) small-intestine classes as shown in \figref{fig:tree}a. Next, we propose a $4$-$2$-$2$-$2$-$2$-$4$ refinement setup, where models first learn $4$ base classes, then progressively add $2$ novel classes at four steps, and finally learn $4$ new classes in the final step.
\figref{fig:tree}b illustrates these settings with a toy example of $3$-$2$-$4$ refinements.}

{\parskip=3pt\noindent\textbf{MM-OR dataset}: The MM-OR dataset~\cite{ozsoy2024mmor} entails multi-modal recordings of $39$ robotic knee replacement surgeries. We follow the official dataset splits for the Azure RGB-D cameras, comprising \num{33298} training, \num{13191} validation, and \num{12055} test images~\cite{ozsoy2024mmor}. We define incremental scenarios using the $18$ most prominent labels: a disjoint setup with $11$-$1$ ($8$ tasks) and a refinement setup following the $11$-$4$-$3$-$2$-$3$ split.}

{\parskip=3pt\noindent\textbf{Syn-Mediverse}: The dataset contains synthetic hospital images from $13$ distinct medical rooms generated with NVIDIA Isaac Sim~\cite{mohan23syn}. The original benchmark comprises $21$ semantic categories, with the dataset split into \num{9000} training, \num{2000} validation, and \num{5000} test images. To create a CISS setting with rich class granularity, we create a finer label-taxonomy covering $144$ classes. First, we filter the $200$ raw object categories retrieved from the simulation by merging classes with duplicate semantics and removing classes without clear descriptors. Further, we ignore classes with fewer than $50$ instances by merging them into broader semantic categories or grouping them by functionality (e.g., clamp and cutting tools). We release the fine-grained Syn-Mediverse training and validation label set at \url{http://topics.cs.uni-freiburg.de}, and the corresponding test labels is hosted as an online benchmark. We define the disjoint $55$-$44$ ($3$ tasks) set-up and incrementally refine medical tools and storage fixtures in our $83$-$11$-$12$-$11$-$13$-$9$-$9$ setting.}

\subsection{Methodology}\label{sec:network}

We propose \net, a variant of the recently introduced TOPICS~\cite{hindel25} framework, tailored for robotic surgery environments. Our approach builds on a DeepLabV3 model with a ResNet-101 backbone and leverages a structured class taxonomy and implicit relations between prior classes to avoid catastrophic forgetting in incremental learning. The class hierarchy is explicitly encoded in the network's final features, which are projected in hyperbolic space. This geometry ensures that classes are equidistant to each other, making it well-suited for representing hierarchical relationships.
During incremental steps, \net~leverages the old model's weights to create hierarchical pseudo-labels for old classes. In addition to the original taxonomy, scarcity, and relation regularization losses employed in TOPICS, \net~introduces modifications to the loss formulation to better align with domain-specific challenges in surgical scenes, while still preserving both explicit and implicit class relationships across learning stages.

\noindent\textbf{Hyperbolic Semantic Segmentation}: 
The last neural network layer operates in hyperbolic space with a constant negative curvature $c$. We leverage the Poincaré ball model, which is a stereographic projection of the upper sheet of a two-sheeted hyperboloid as shown in~\figref{fig:pc}. Multinomial regression represents each class $y$ as a hyperplane in the Poincaré ball characterized by an offset $o_y \in \mathbb{D}_c^N$ from the center and an orientation $r_y \in {T}\mathbb{D}_c^N$. The likelihood of class $y$ is based on the signed distance to its hyperplane.





\noindent\textbf{Hierarchical Segmentation}: 
We enforce hyperplanes on the Poincaré ball to reflect the hierarchy of semantic classes. Therefore, leaf nodes and all their ancestors are modeled as separate output classes $\mathcal{V}$ and a combination of ancestor $\mathcal{A}$ and descendant $\mathcal{D}$ logits ($s$) is employed in the loss function. \net~defines the hierarchical dice loss according to: 
\begin{align*}
\mathcal{L} = \alpha \bigg( \sum_{v \in \mathcal{V}} -l \log \Big( \min_{u \in \mathcal{A}_v} (s_u) \Big ) - (1-l) \log \Big( 1-\max_{u \in \mathcal{D}_v} (s_u) \Big ) \bigg) \\ + \beta \mathcal{L}_D\Big( \max_{u \in \mathcal{D}_v} 
(s_u), l \Big )
+ \gamma \mathcal{L}_{CE}\Big( \max_{u \in \mathcal{D}_v} 
(s_u), l \Big ),
\end{align*}
which integrates the dense Dice loss $\mathcal{L}_D$~\cite{milletari2016v} at various hierarchical levels to address the class imbalance and penalize high prediction scores of sibling classes.

\noindent\textbf{Hierarchical Pseudo-Labeling}: 
TOPICS employs old model's weights to create pseudo-labels for old classes. Thereby, leaf-classes with a sigmoid score above a uniform threshold ($s$=$0.5$) are adopted. We observed that this conventional pseudo-labeling is insufficient for surgical environments characterized by diverse backgrounds, such as those found in the MM-OR dataset. To address this problem, \net~introduces hierarchical pseudo-labeling. Specifically, we define thresholds $s_i$ for each of the $i$ hierarchical levels and assign the most specific (descendant) class whose confidence exceeds the corresponding $s_i$.

\noindent\textbf{Creating Hierarchical Graph}: 
To avoid catastrophic forgetting in incremental learning, we construct deep hierarchies for each dataset. Given the initial set of label classes in the dataset, we use OpenAI's GPT-4o~\cite{openai2024gpt4technicalreport} to create semantically meaningful class hierarchies as depicted in \figref{fig:tree} using the following prompt "\textit{Here are the list of classes for <dataset>. Create a hierarchy of classes depending on visual similarity.}" We also investigate graphs linking objects that frequently appear in close proximity in \secref{sec:quan}.

\section{Experimental Evaluation}


\subsection{Experimental Setup} \label{sec:setup}
We train \net~for $60$ epochs on Endovis18/MM-OR and $30$ epochs on Syn-Mediverse. We use a polynomial scheduler with initial LR of $0.03$ for base training and $0.01$ in all incremental steps ($0.02$ for Syn-Mediverse). The hierarchical dice loss uses $\alpha=5$, $\beta=0.7$ and $\gamma=0.3$. We set the curvature $c$ to $3$ and fix the hierarchical pseudo-labeling thresholds $s_0$, $s_1$ and $s_2$ to $0.6$, $0.6$ and $0.4$, which were chosen empirically based on validation performance. We apply horizontal flipping and non-empty cropping: (768, 768) for MM-OR, (512, 1024) for Syn-Mediverse, and (512, 640) for Endovis18. For Endovis18, we add affine, elastic, contrast, and brightness augmentations. Further settings are adapted from \cite{hindel25}.

\subsection{Quantitative Results}\label{sec:quan}
We compare \net~with TOPICS~\cite{hindel25}, PLOP~\cite{Douillard2020PLOPLW}, MiB~\cite{cermelli2020mib}, MiB+AWT~\cite{goswami2023attribution} and DKD~\cite{baek2022_dkd} using their published code, identical augmentations (\secref{sec:setup}), and model- and dataset-specific LR tuning.
PLOP~\cite{Douillard2020PLOPLW} requires square inputs and is trained on $512\times512$ crops for Syn-Mediverse and EndoVis18.
All models have $5.8$M parameters and are evaluated on an NVIDIA RTX A6000 GPU. Inference takes $118$ms for MIB/AWT/PLOP, $130$ms for DKD, and $145$ms for TOPICS/TOPICS+.
We report the mean intersection-over-union (mIoU) metric on base ($\mathcal{C}_1$), novel ($\mathcal{C}_{2:T}$) and all classes ($\mathcal{C}_{1:T}$).

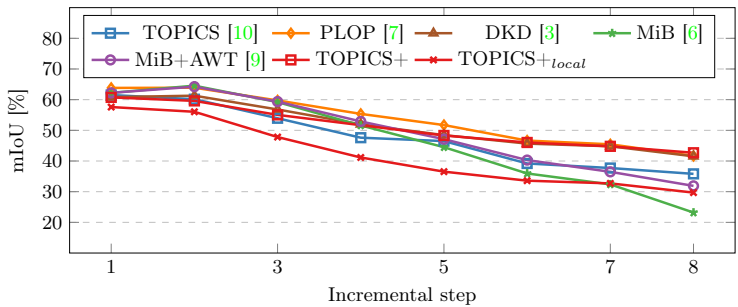
\begin{figure}[!htbp]
        \centering
        \resizebox{0.8\columnwidth}{!}{%
        \begin{tikzpicture} [font=\small]
        \begin{axis}[
            title={},
            ylabel={mIoU [\%]},
            legend style={font=\footnotesize},
            xlabel={Incremental step},
            xmin=0.5, xmax=8.5,
            ymin=10, ymax=90,
            ytick={20,30,40,50, 60, 70, 80},
            xtick={1,3,5,7,8},
            legend pos=north east,
            legend columns=4, 
            ymajorgrids=true,
            grid style=dashed,
            height=5.5cm
        ]

        \addplot[
            color=color_blue,
            mark=square,
            line width=0.4mm,
            ]
            coordinates {
            (1,61.51)(2,60.15)(3,53.94)(4,47.60)(5,46.49)(6,39.20)(7,37.71)(8,35.81)
            };
        \addlegendentry{TOPICS~\cite{hindel25}}
        \addplot[
            color=color_orange,
            mark=diamond,
            line width=0.4mm,
            ]
            coordinates {
            (1,63.83)(2,63.89)(3,59.75)(4,55.36)(5,51.72)(6,46.66)(7,45.45)(8,41.67)
            };
        \addlegendentry{PLOP~\cite{Douillard2020PLOPLW}}
        \addplot[
            color=color_brown,
            mark=triangle,
            line width=0.4mm,
            ]
            coordinates {
            (1,60.82)(2,61.30)(3,56.80)(4,51.96)(5,48.44)(6,45.67)(7,44.80)(8,41.53)
            };
        \addlegendentry{DKD~\cite{baek2022_dkd}}
                \addplot[
            color=color_green,
            mark=star,
            line width=0.4mm,
            ]
            coordinates {
            (1,62.22)(2,64.56)(3,59.20)(4,51.60)(5,44.51)(6,35.94)(7,32.38)(8,23.20)
            };
        \addlegendentry{MiB\cite{cermelli2020mib}}
        \addplot[
            color=color_purple,
            mark=o,
            line width=0.4mm,
            ]
            coordinates {
            (1,62.22)(2,64.24)(3,59.36)(4,52.90)(5,47.12)(6,40.30)(7,36.47)(8,31.89)
            };
        \addlegendentry{MiB+AWT~\cite{goswami2023attribution}}

        \addplot[
            color=color_red,
            mark=square,
            line width=0.4mm,
            ]
            coordinates {
            (1,60.74)(2,59.60)(3,55.04)(450.61)(5,48.32)(6,46.04)(7,44.79)(8,42.69)
            };
        \addlegendentry{\net}
        \addplot[
            color=color_red,
            mark=x,
            line width=0.4mm,
            ]
            coordinates {
            (1,57.59)(2,56.05)(3,47.81)(4,41.15)(5,36.53)(6,33.59)(7,32.70)(8,29.70)
            };
        \addlegendentry{\net$_{local}$}
        
    \end{axis}
    \end{tikzpicture}}
\caption{Performance at every increment on MM-OR 11-1 (8 tasks) setting.}
\label{fig:miou10tasks}
\vspace{-0.3cm}
\end{figure}

We present the results on disjoint increments (defined in \secref{sec:dataset}) in \tabref{tab:icl}. We observe that \net~outperforms TOPICS in all three evaluated scenarios. This underscores the benefit of our hierarchical dice loss for robotic surgical environments. The methods MiB+AWT and DKD excel on novel classes but struggle with base classes, leading to a lower overall score. 
We show in \figref{fig:miou10tasks} that \net~outperforms PLOP even if it has $3$pp lower performance at t=$1$ which highlights its high potential to maintain prior knowledge.
Additionally, TOPICS exhibits false-negative background predictions, likely due to the close proximity of labeled instruments and background objects. Our hierarchical pseudo-labeling approach effectively mitigates this issue as evident in a higher score on novel classes.
Additionally, we show that using a location-based hierarchy (\net~$_{local}$) significantly reduces the retained knowledge of base classes and generalization on new classes in \figref{fig:miou10tasks}.

\begin{table*}[!htbp]
\centering
\scriptsize
\setlength{\tabcolsep}{4pt}
\caption{Continual semantic segmentation results on disjoint incremental settings in mIoU (\%). Tasks defined as $\mathcal{C}^1$-$\mathcal{C}^T$($T$ tasks).}
\label{tab:icl}
\begin{tabular}
{l|ccc|ccc|ccc}
 \toprule
  & \multicolumn{3}{c|}{Endovis18} & \multicolumn{3}{c|}{MM-OR} & \multicolumn{3}{c}{Syn-Mediverse} \\
 & \multicolumn{3}{c|}{\textbf{8-1 (4 tasks)}} 
 & \multicolumn{3}{c|}{\textbf{11-1 (8 tasks)}} &
 \multicolumn{3}{c}{{\textbf{55-44 (3 tasks)}}} \\
  \cmidrule{2-10}
\textbf{Method} & 1-8 & 8-12 & all & 1-11 & 11-18 & all & 1-55 & 55-144 & all \\
\midrule
PLOP~\cite{Douillard2020PLOPLW} 
& 36.12 & 43.14 & 41.43 &
 49.17 & 23.53 & 41.67 &
 50.98 & 16.10 & 29.27 \\
MiB~\cite{cermelli2020mib} & 41.28 & 49.97 & 47.05 &
17.85 & 24.63 & 23.20 &
 50.51 & 44.06 & 46.49 \\
MiB + AWT~\cite{goswami2023attribution} & 43.09 & 51.06 & 48.56 &
31.55 & 25.61 & 31.89 &
51.26 & \textbf{48.12} & 49.31 \\
DKD~\cite{baek2022_dkd} & 34.32 & 40.46 & 37.93 &
 44.64 & \textbf{30.38} & 41.53 &
 50.53 & 29.91 & 37.69 \\
TOPICS~\cite{hindel25} & 
\textbf{49.09} & 48.20 & 48.77 &
 46.15 & 20.88 & 35.81 &
 57.82 & 44.91 & 49.79 \\
\midrule
\net~(Ours) & 47.38 & \textbf{51.96} & \textbf{49.05} &
 \textbf{51.48} & 26.56 & \textbf{42.69} &
 \textbf{59.83} & 44.30 & \textbf{50.16} \\
\hline
\end{tabular}
\end{table*}

\begin{table*}[t]
\centering
\scriptsize
\setlength{\tabcolsep}{4pt}
\caption{Continual semantic segmentation results on hierarchical refinements in mIoU (\%). Tasks defined as $\mathcal{C}^1$-$\mathcal{C}^2$- ... -$\mathcal{C}^T$.}
\label{tab:hicl}
\begin{tabular}
{l|ccc|ccc|ccc}
 \toprule
  & \multicolumn{3}{c|}{Endovis18} & \multicolumn{3}{c|}{MM-OR} & \multicolumn{3}{c}{Syn-Mediverse} \\
 & \multicolumn{3}{c|}{\textbf{4-2-2-2-2-4}} 
 & \multicolumn{3}{c|}{\textbf{11-4-3-2-3}} &
 \multicolumn{3}{c}{\textbf{83-11-12-11-13-9-9}} \\
  \cmidrule{2-10}
\textbf{Method} & 1-4 & 5-16 & all & 1-11 & 11-22 & all & 1-83 & 84-148 & all \\
\midrule
PLOP~\cite{Douillard2020PLOPLW} 
& 44.28 & 15.40 & 28.05 &
 40.63 & 32.78 & 38.40 &
 39.73 & 15.02 & 28.50 \\
MiB~\cite{cermelli2020mib} & 35.38 & 26.95 & 32.80 &
 50.54 & 28.19 & 39.84 &
 41.03 & 37.15 & 39.27 \\
MiB + AWT~\cite{goswami2023attribution} & 36.96 & 26.17 & 32.69 &
 \textbf{50.59} & 32.32 & 42.21 &
 41.13 & \textbf{44.18} & 42.52 \\
DKD~\cite{baek2022_dkd} & 38.36 & 1.61 & 16.29 &
47.92 & 7.06 & 26.69 &
 50.41 & 6.34 & 30.38 \\
TOPICS~\cite{hindel25} & 
 \textbf{44.72} & 27.61 & 33.83 &
 50.38 & 31.88 & 40.12 &
 \textbf{54.58} & 39.83 & 47.88 \\
\midrule
\net~(Ours) & 44.12 & \textbf{29.99} & \textbf{35.13} & 
 50.20 & \textbf{38.52} & \textbf{45.10} &
 53.27 & 42.13 & \textbf{48.20} \\
\hline
\end{tabular}
\end{table*}

Next, we present the results on label refinements according to \secref{sec:dataset} in \tabref{tab:hicl}. We note that \net~achieves the best balance in maintaining knowledge on base classes while learning refinements in all settings. \net~significantly exceeds TOPICS in terms of novel class performance, which we attribute to our hierarchical dice loss, as it effectively addresses class imbalance.
Further, \net~outperforms all non-hierarchical baselines by $5.68$ pp on the refinement Syn-Mediverse setting which underscores the benefit of hierarchical encoding for retaining knowledge in class-rich incremental settings. Moreover, we find that DKD mitigates forgetting of base classes but is unable to generalize on the refined novel classes which is likely due to this method’s reliance on a frozen backbone, which constrains its learning capacity.


\subsection{Qualitative Results}

\begin{figure*}[t]
\centering
\begin{subfigure}[b]{0.395\textwidth}
\vskip 1pt
\centering
\scriptsize
\setlength{\tabcolsep}{0.05cm}
{
\renewcommand{\arraystretch}{0.2}
\newcolumntype{M}[1]{>{\centering\arraybackslash}m{#1}}
\begin{tabular}{cM{1cm}M{1cm}M{1cm}M{1cm}}
& Input image & TOPICS \cite{hindel25} & Ours & Improv./ Error \\
(i) & \includegraphics[width=\linewidth, frame]{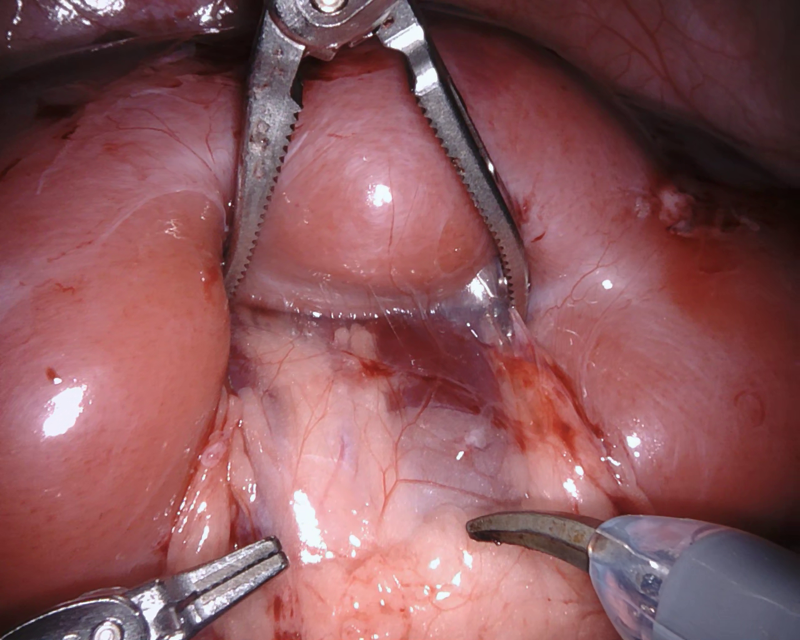} & \includegraphics[width=\linewidth, frame]{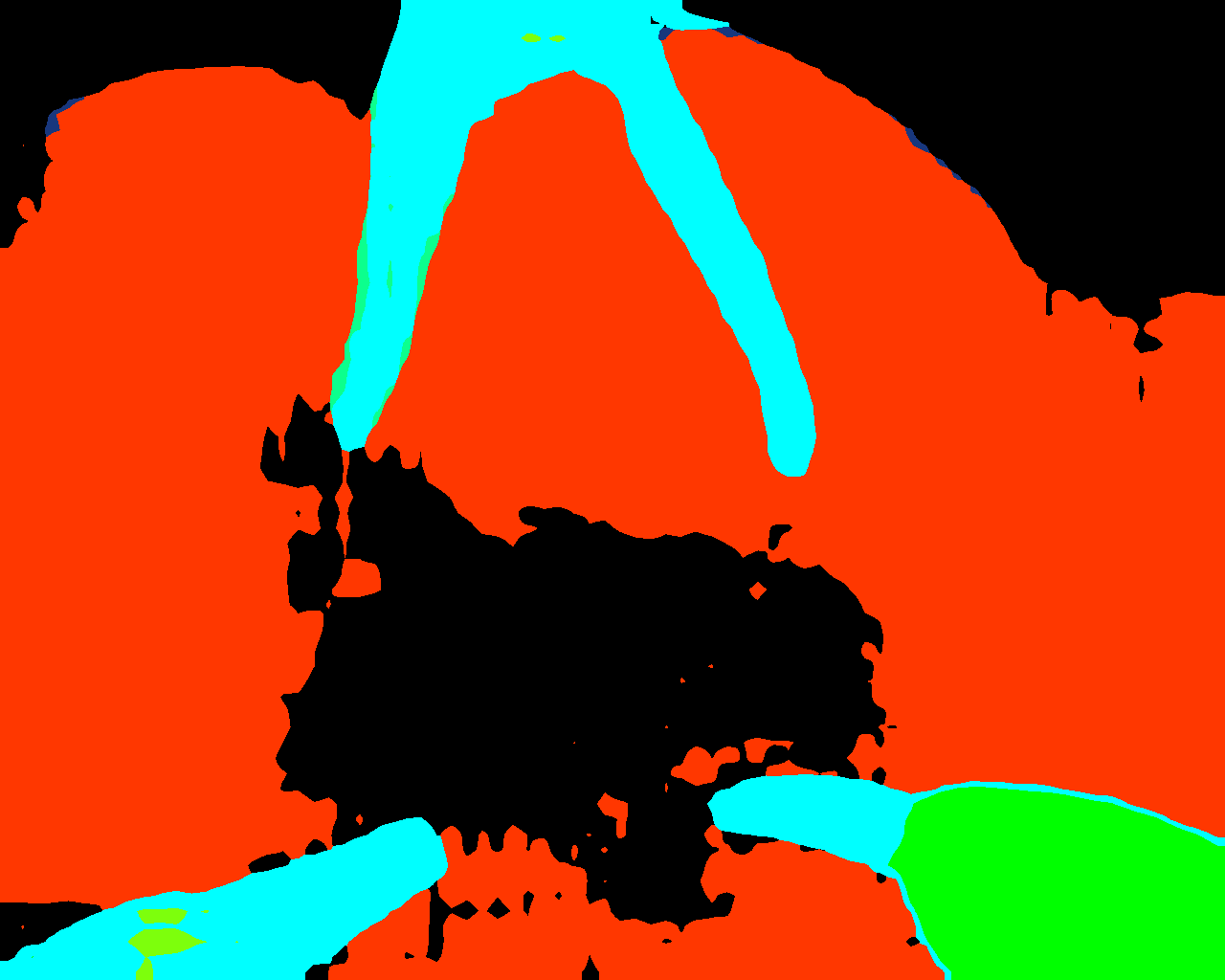} & \includegraphics[width=\linewidth, frame]{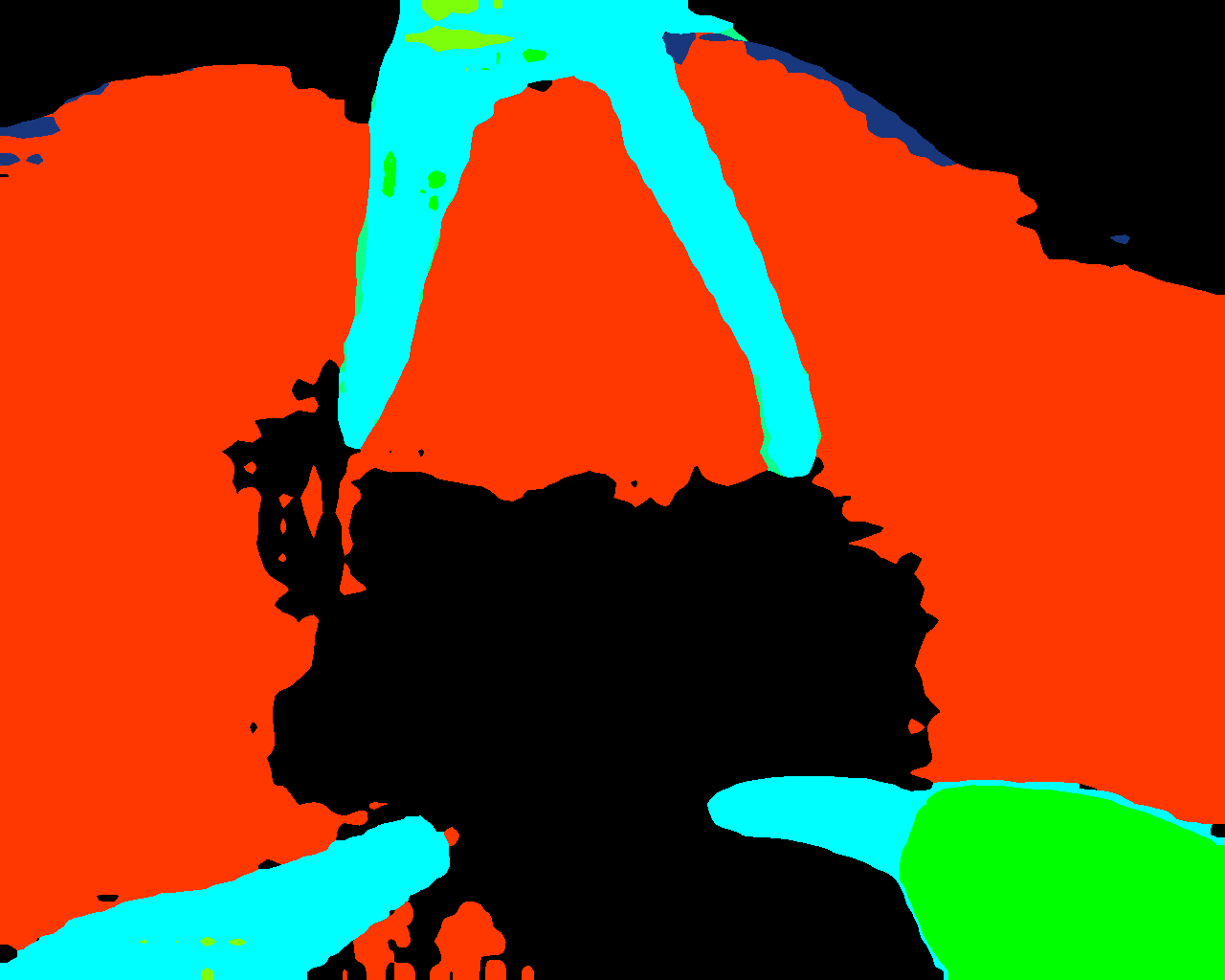} & \includegraphics[width=\linewidth, frame]{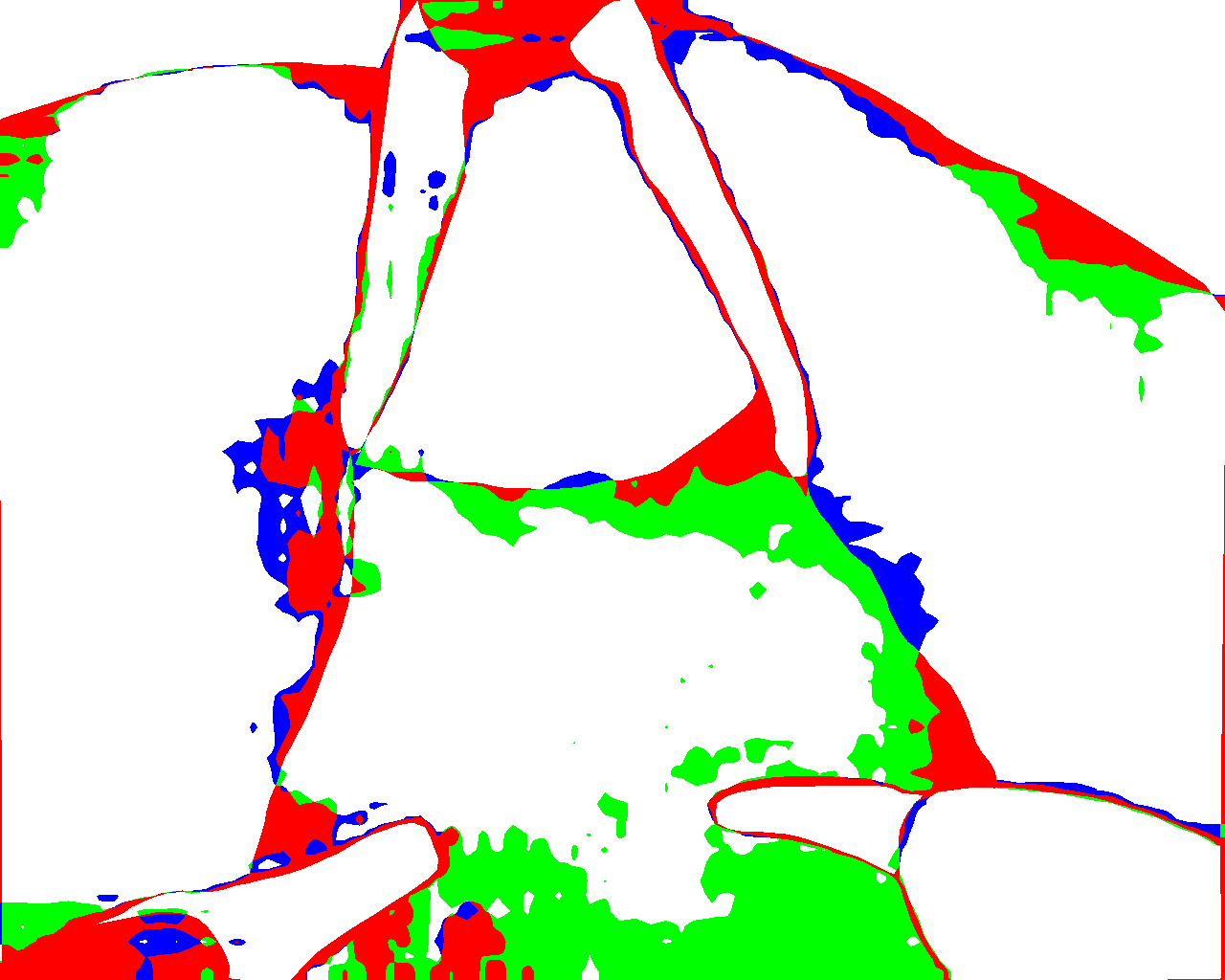}\\
\\
(ii) &  \includegraphics[width=\linewidth, frame]{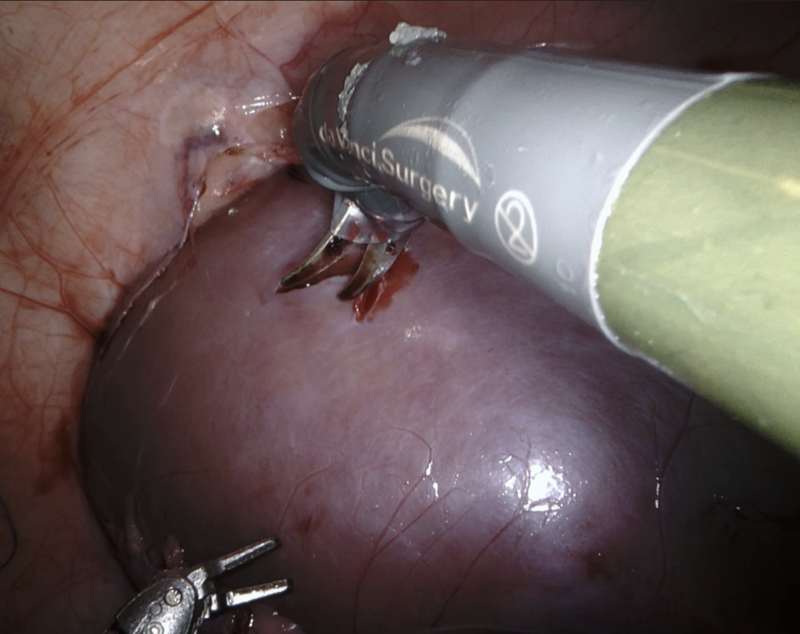} & \includegraphics[width=\linewidth, frame]{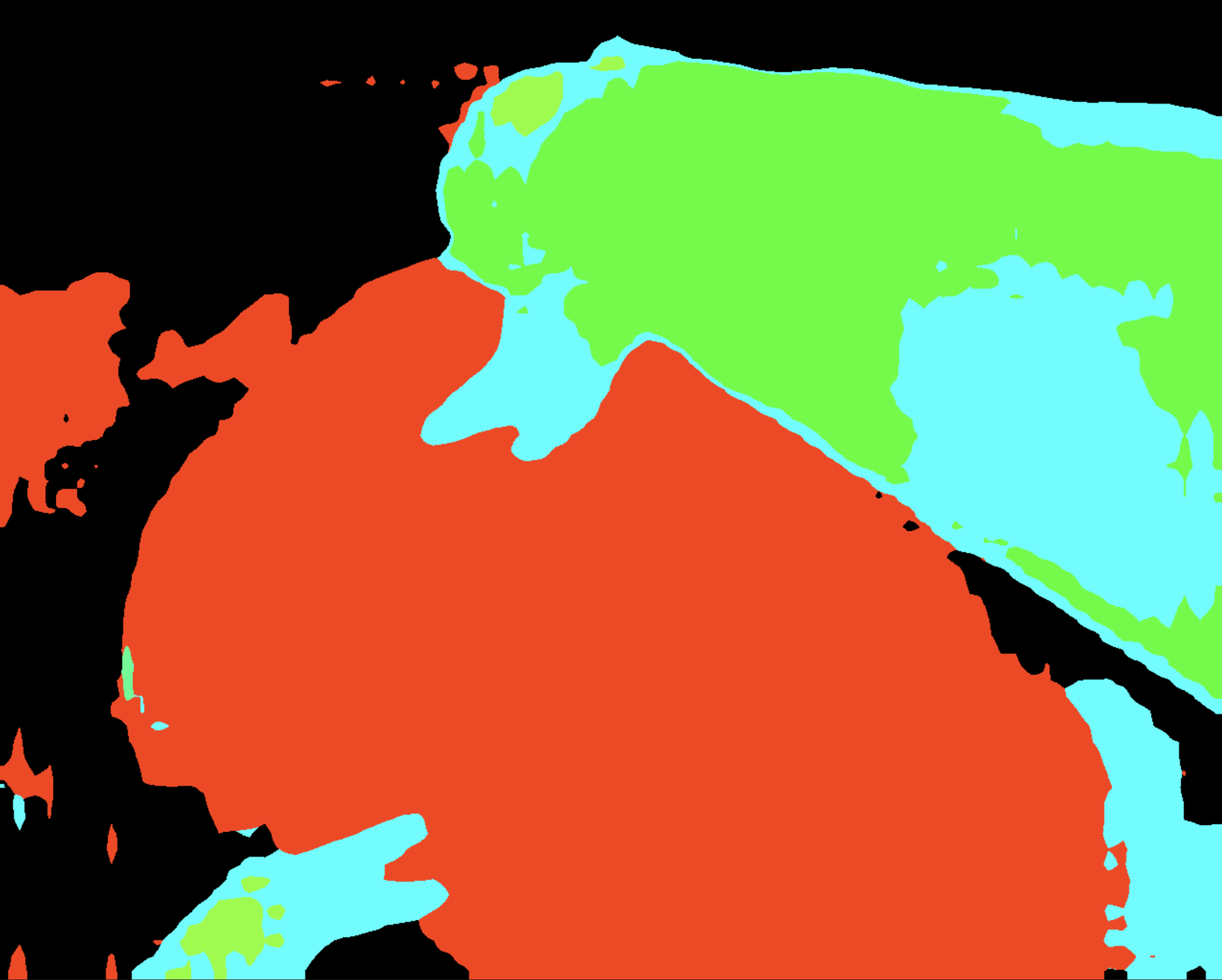} & \includegraphics[width=\linewidth, frame]{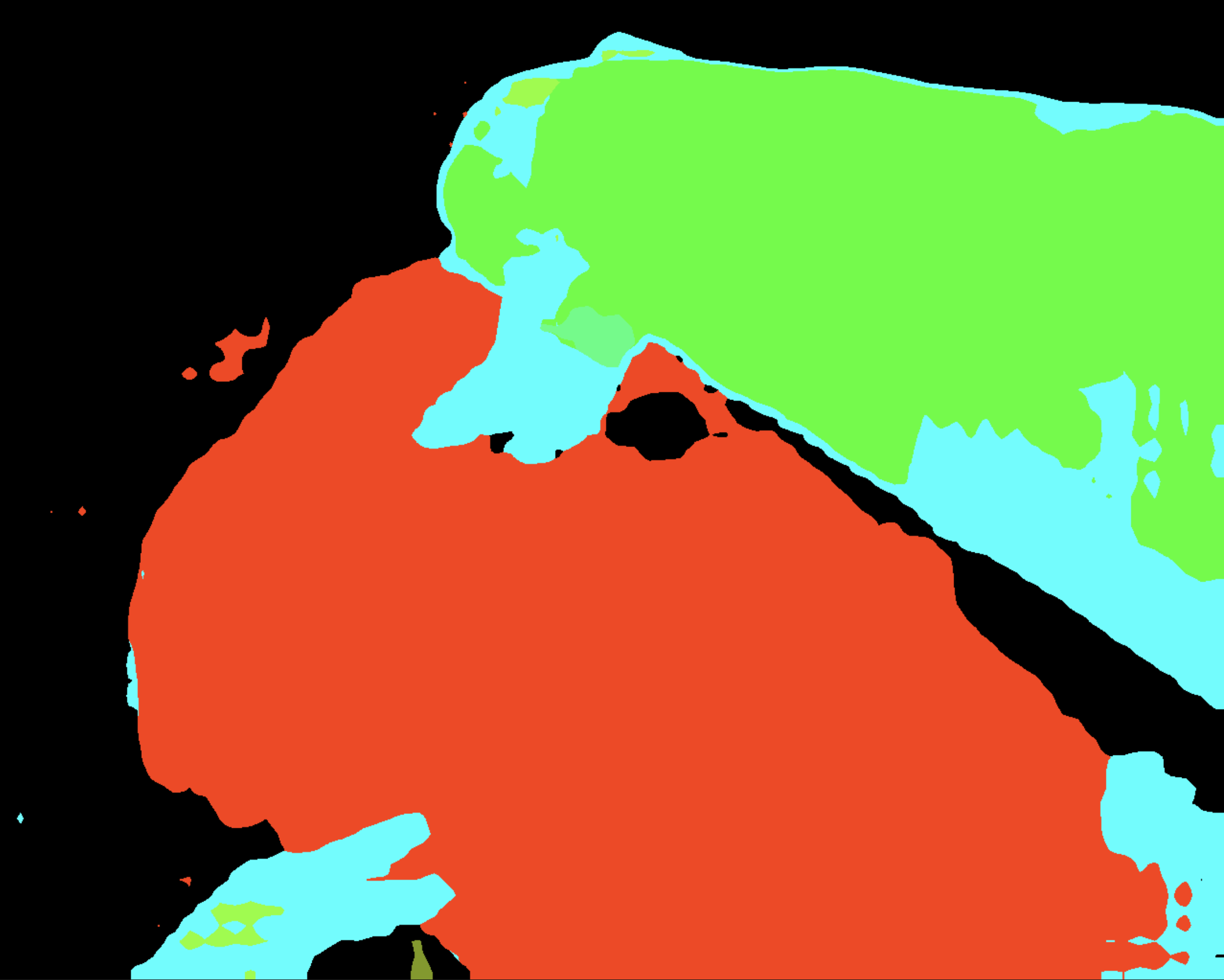} & \includegraphics[width=\linewidth, frame]{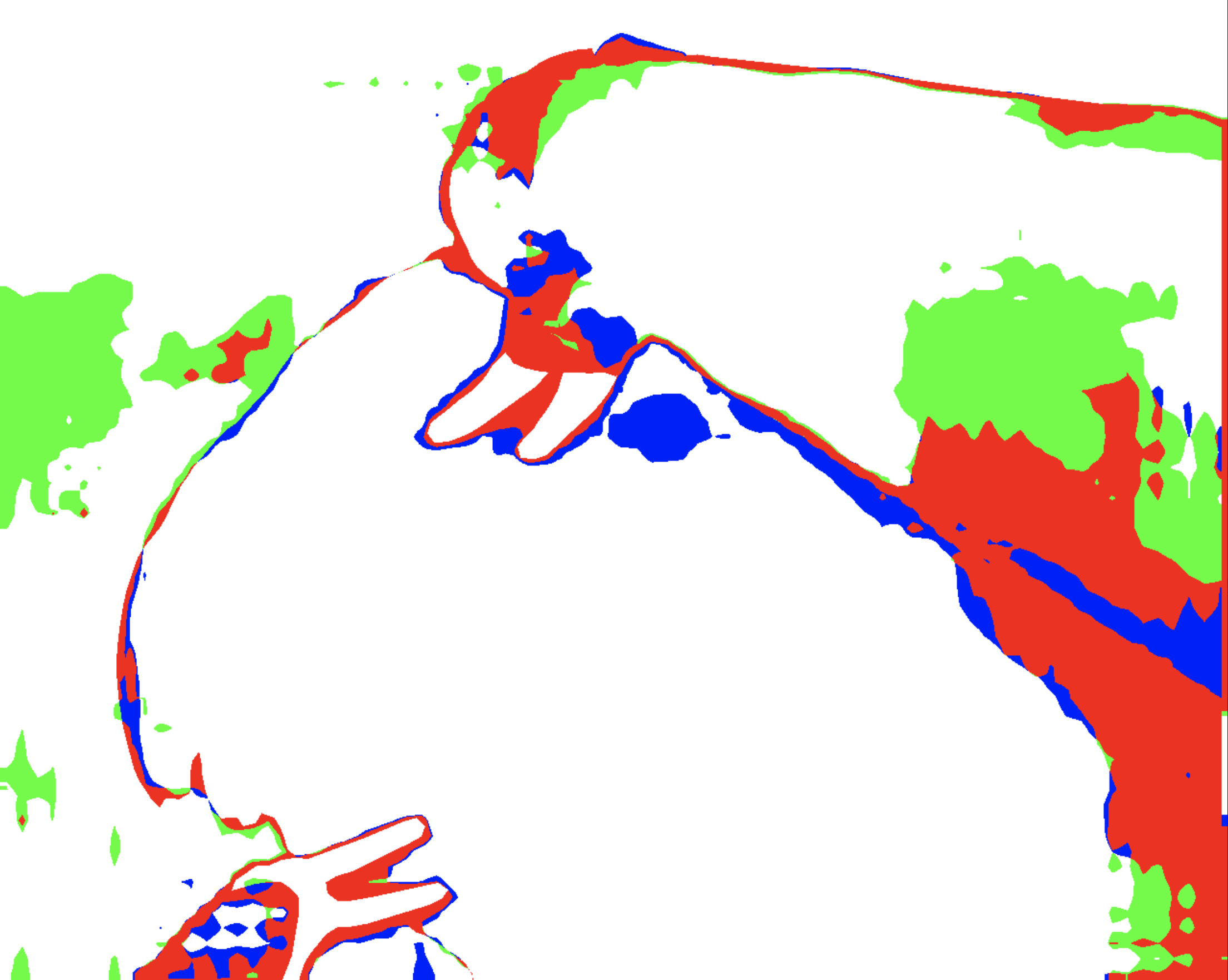}\\
\\
(iii) & \includegraphics[width=\linewidth, frame]{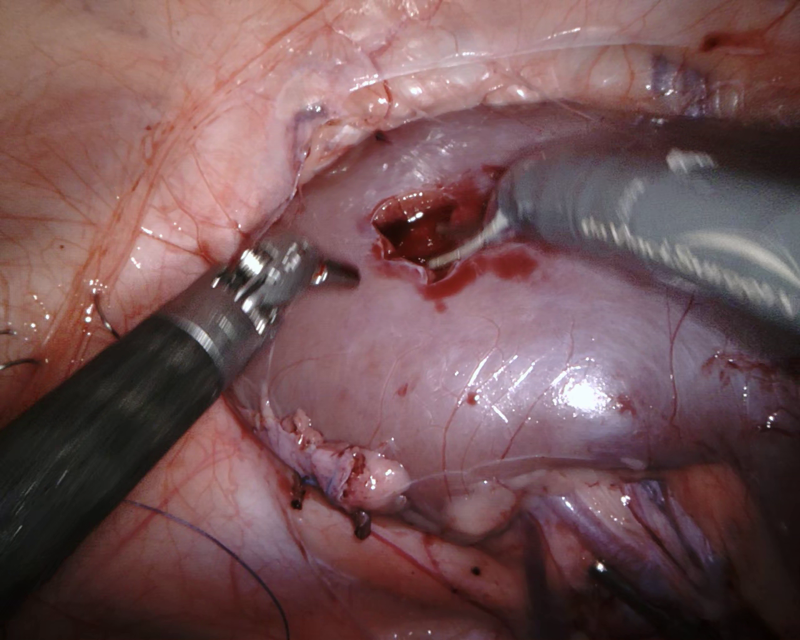} & \includegraphics[width=\linewidth, frame]{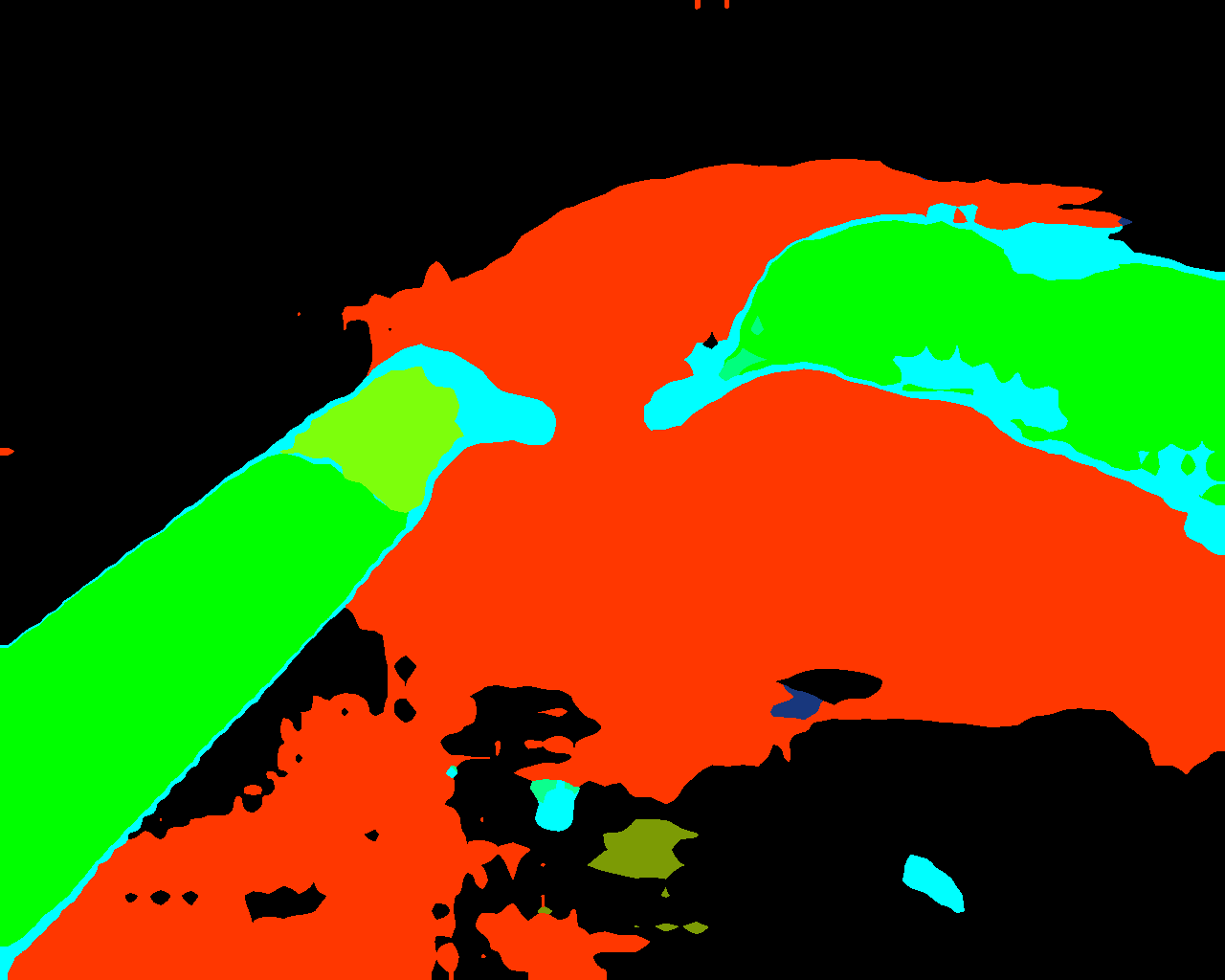} & \includegraphics[width=\linewidth, frame]{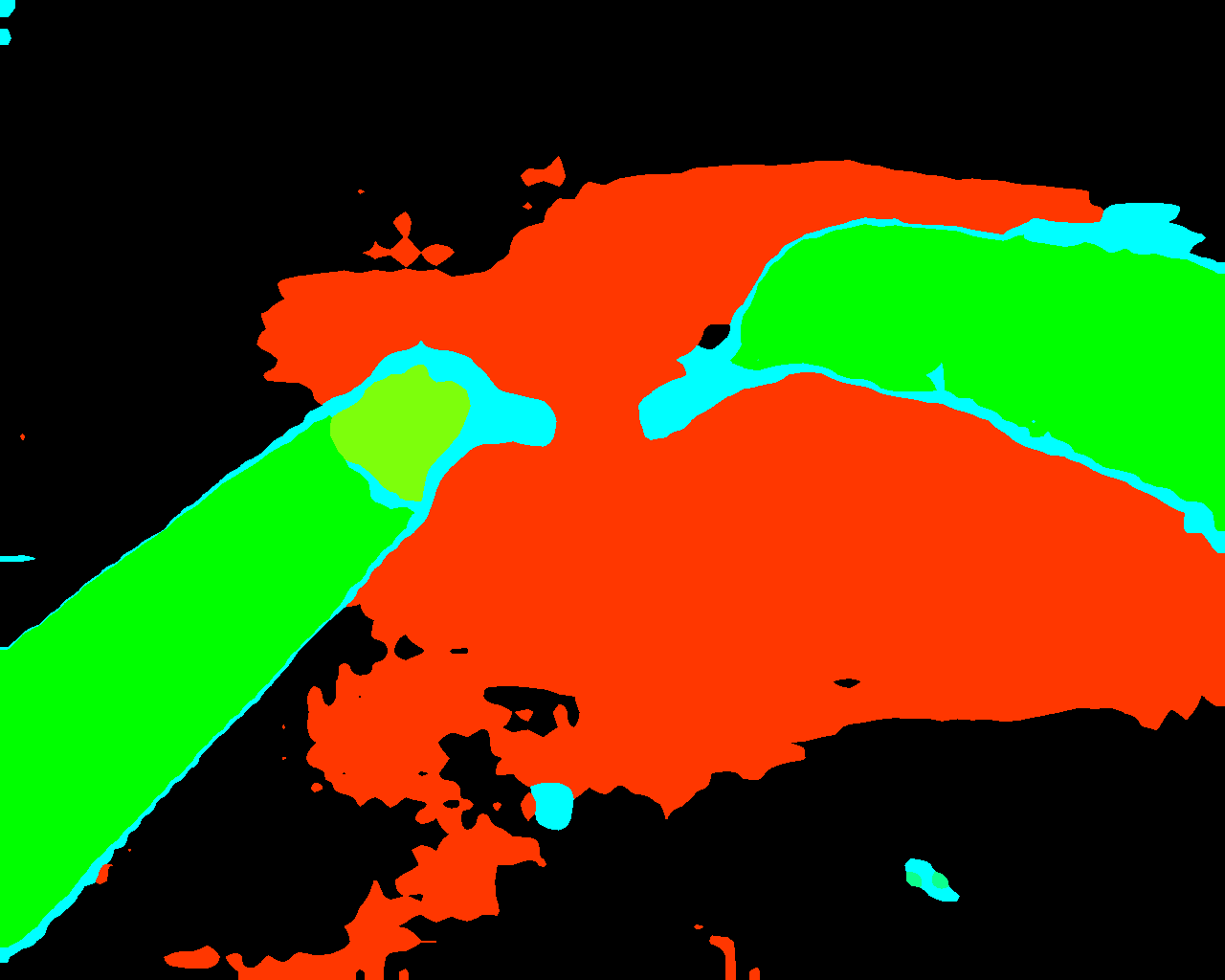} & \includegraphics[width=\linewidth, frame]{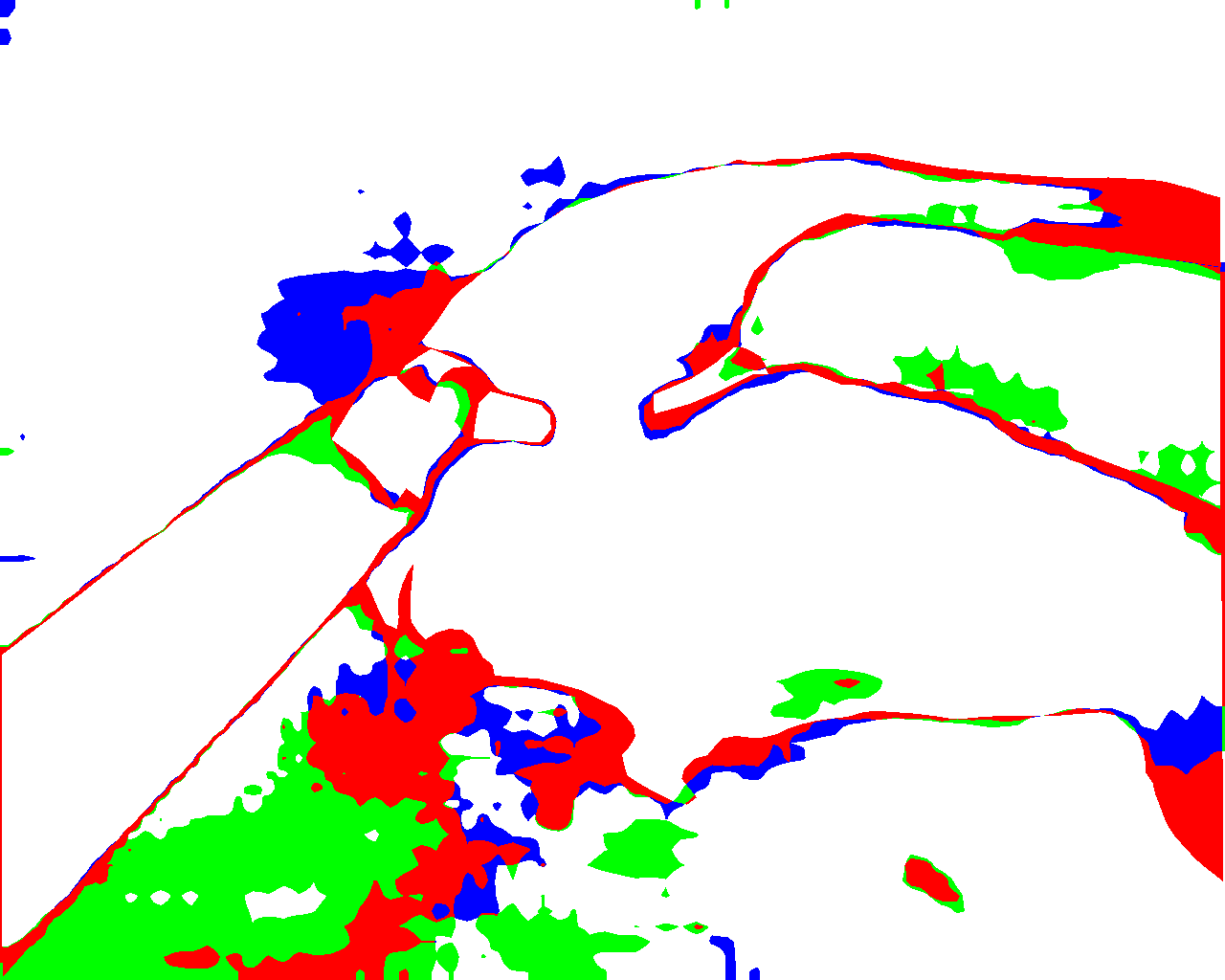}\\
\\
\end{tabular}
}
\caption{Semantic segmentation results on Endovis18 4-2-2-2-2-4.}
\label{fig:qual-analysisIS}
\end{subfigure}
\hfill
\begin{subfigure}[b]{0.595\textwidth}
\vskip 1pt
\centering
\scriptsize
\setlength{\tabcolsep}{0.05cm}
{
\renewcommand{\arraystretch}{0.2}
\newcolumntype{M}[1]{>{\centering\arraybackslash}m{#1}}
\begin{tabular}{cM{1.45cm}M{1.45cm}M{1.45cm}M{1.45cm}}
& Input image & TOPICS \cite{hindel25} & Ours & Improv./ Error \\
(i) & \includegraphics[width=\linewidth, frame]{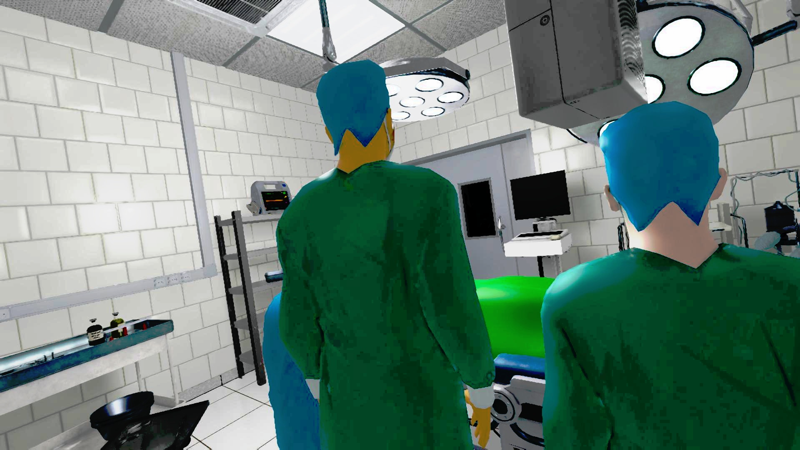} & \includegraphics[width=\linewidth, frame]{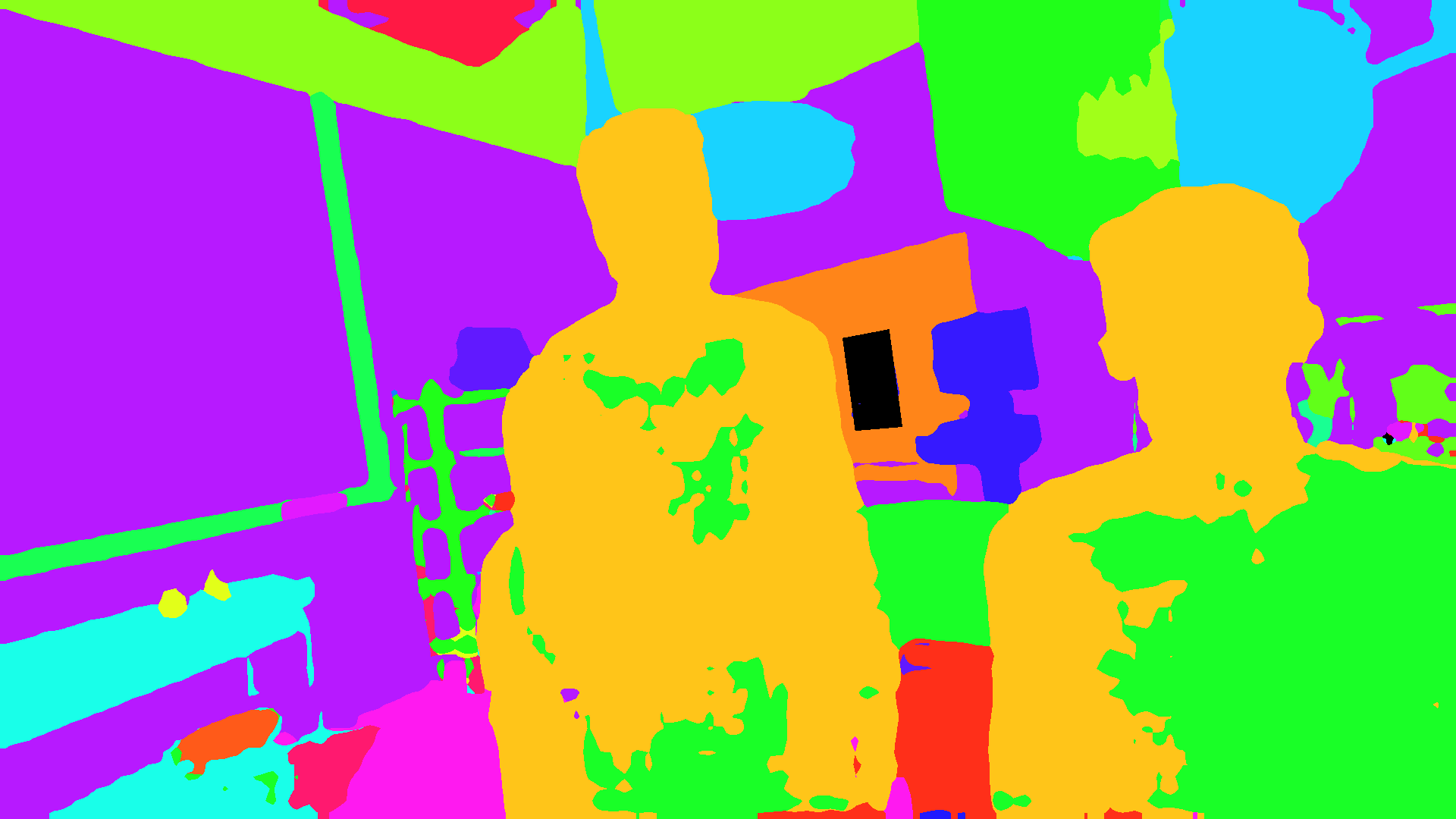} & \includegraphics[width=\linewidth, frame]{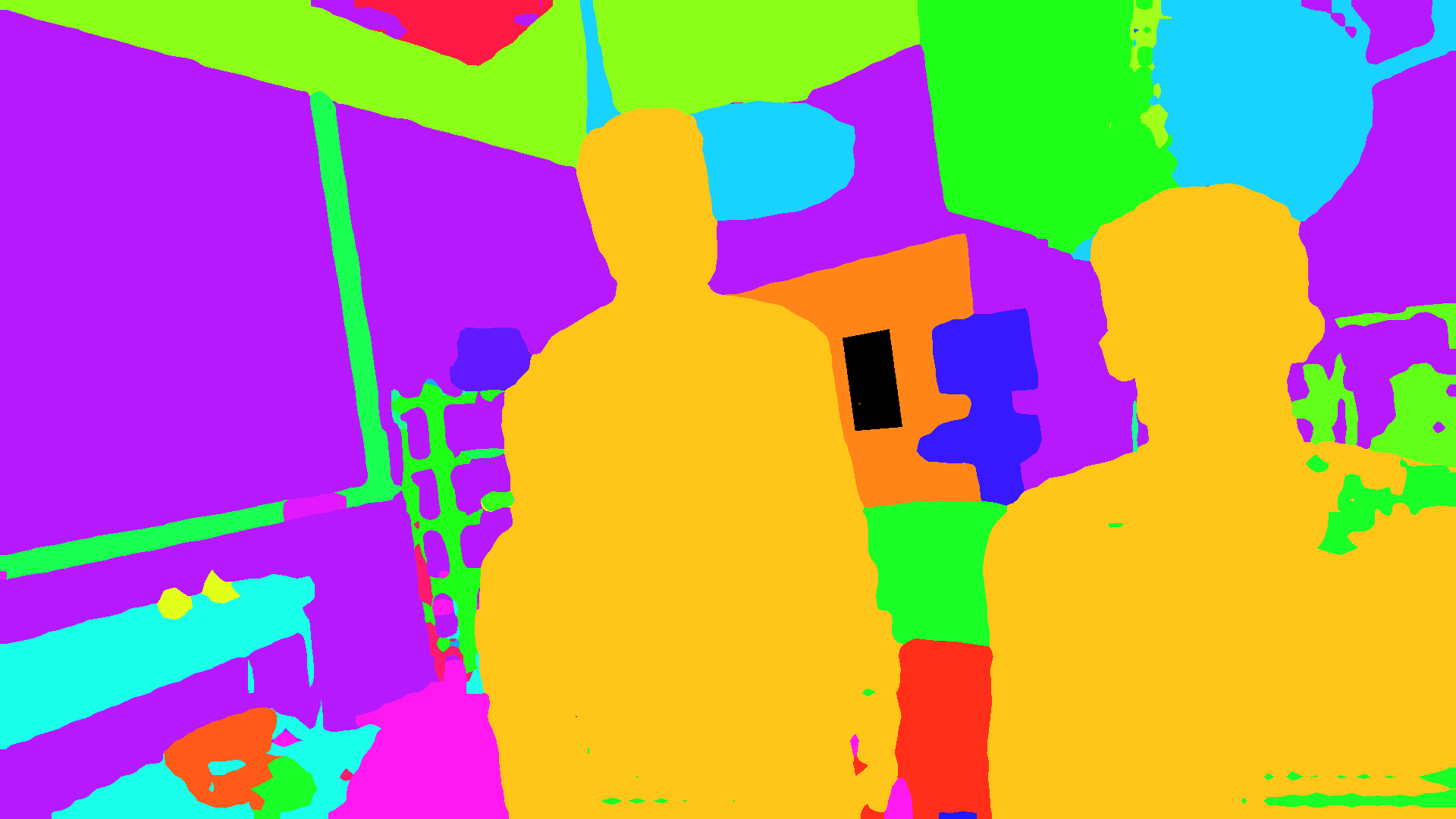} & \includegraphics[width=\linewidth, frame]{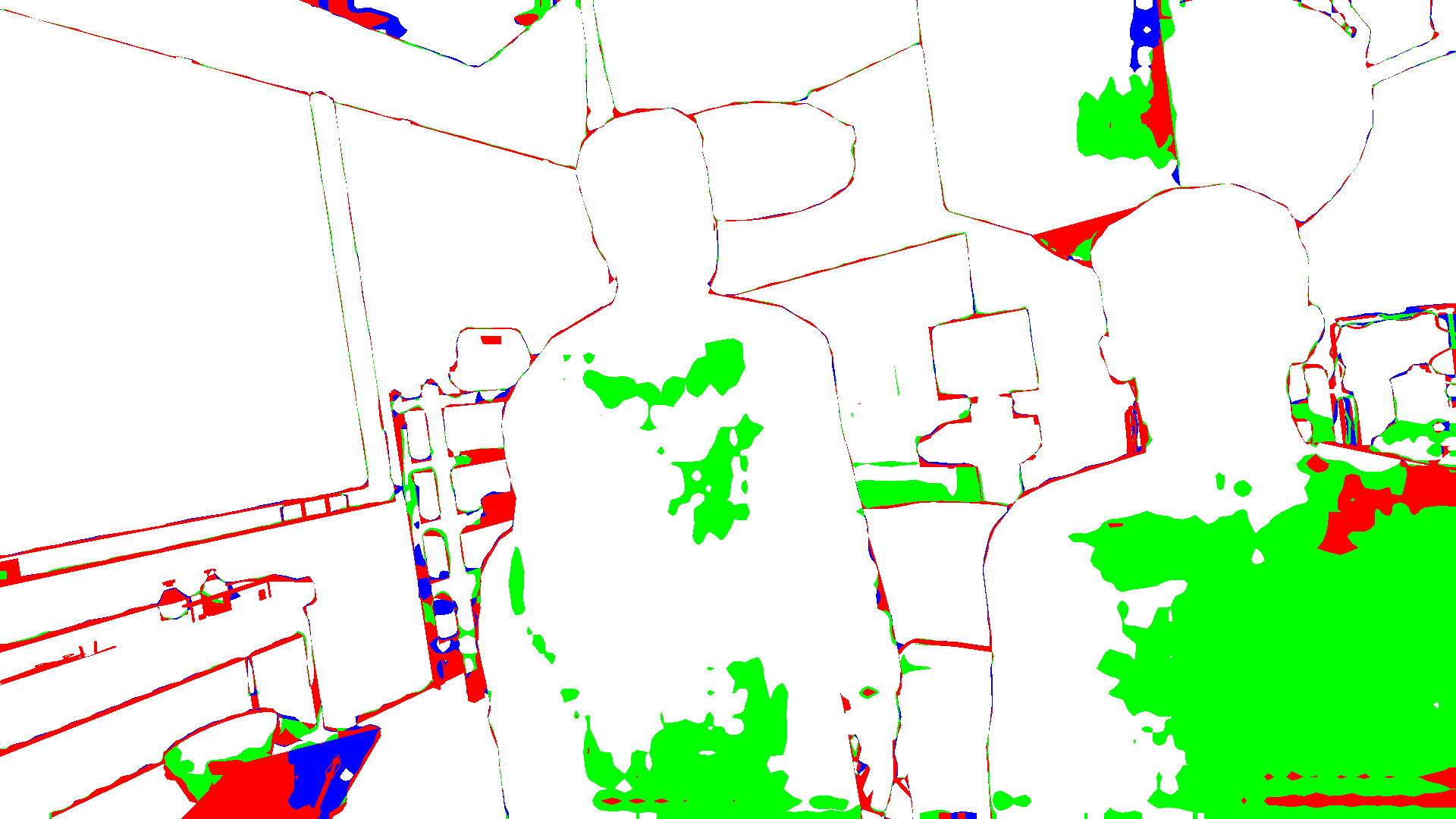}\\
\\
(ii) &  \includegraphics[width=\linewidth, frame]{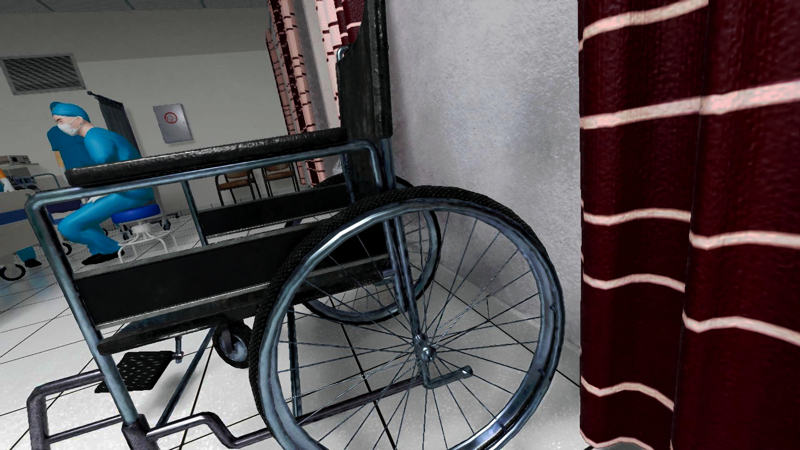} & \includegraphics[width=\linewidth, frame]{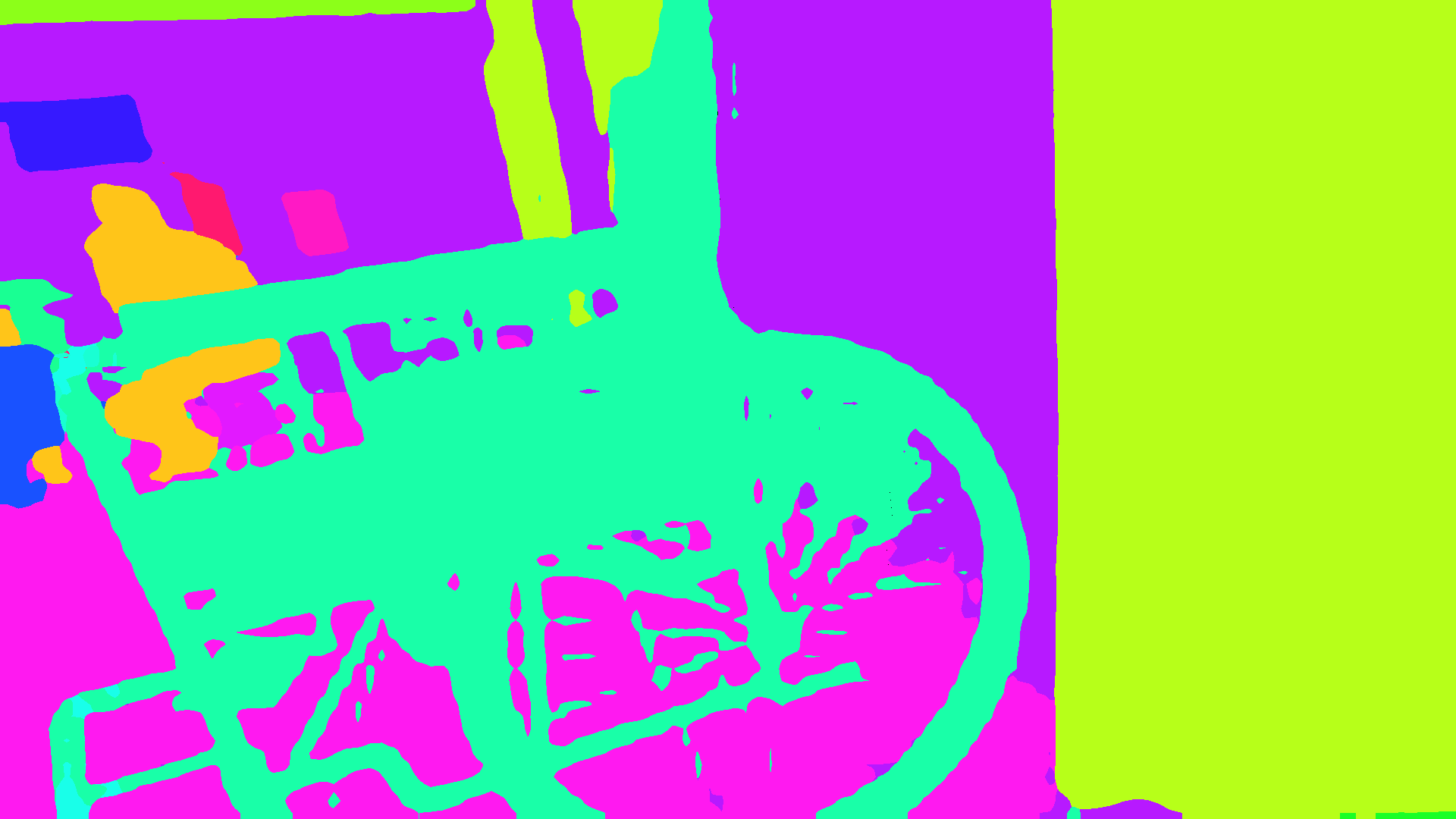} & \includegraphics[width=\linewidth, frame]{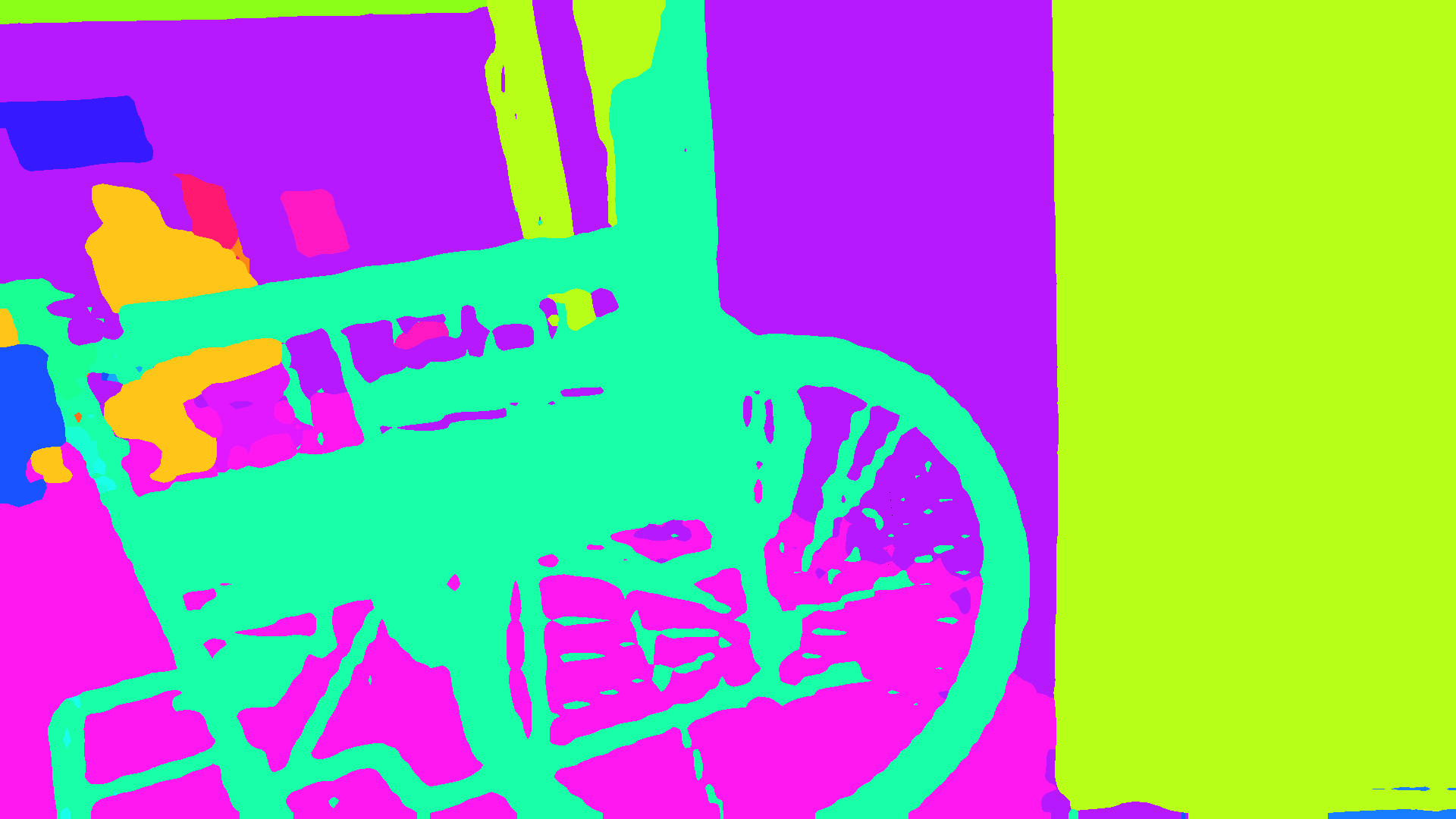} & \includegraphics[width=\linewidth, frame]{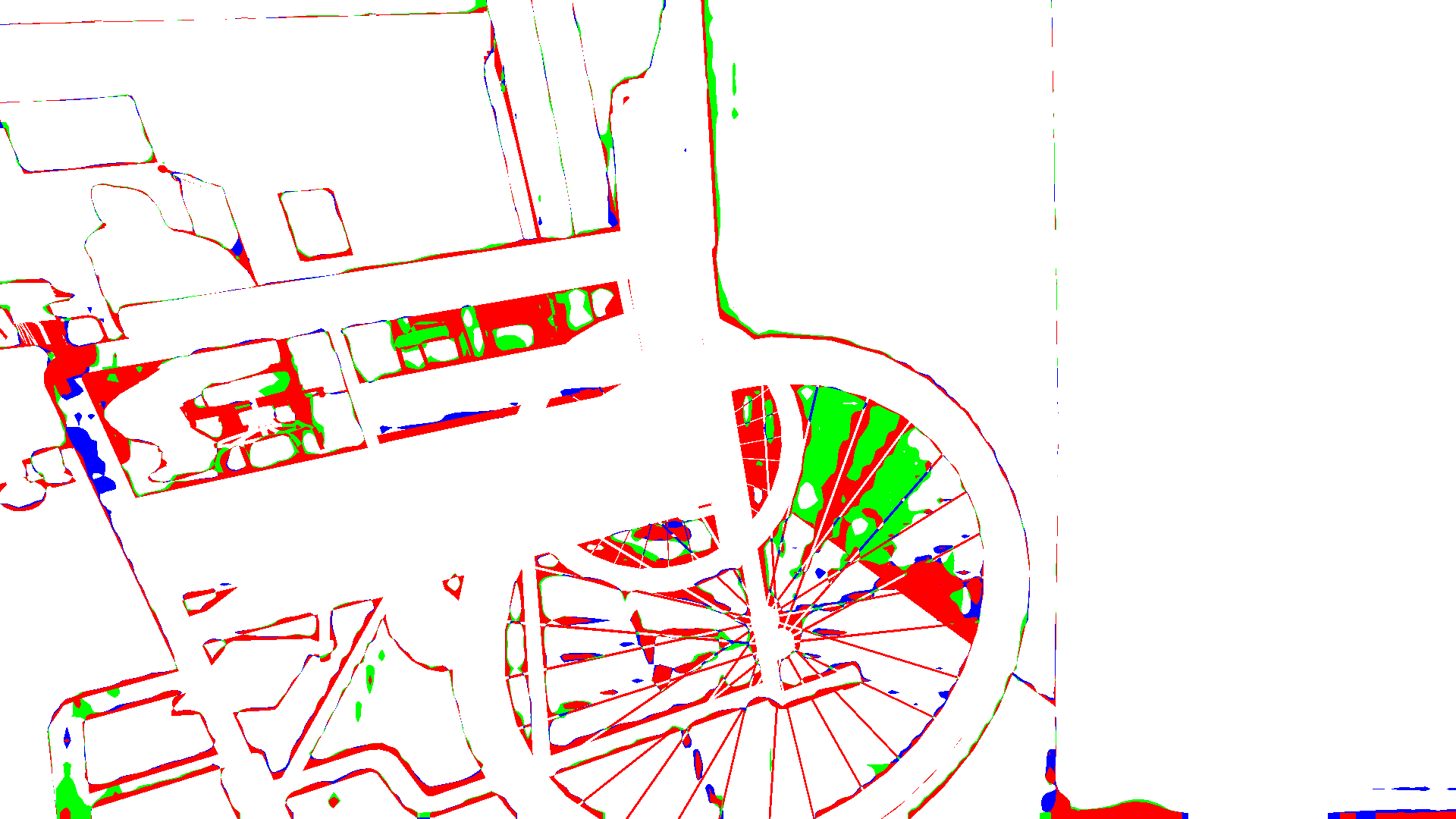}\\
\\
(iii) & \includegraphics[width=\linewidth, frame]{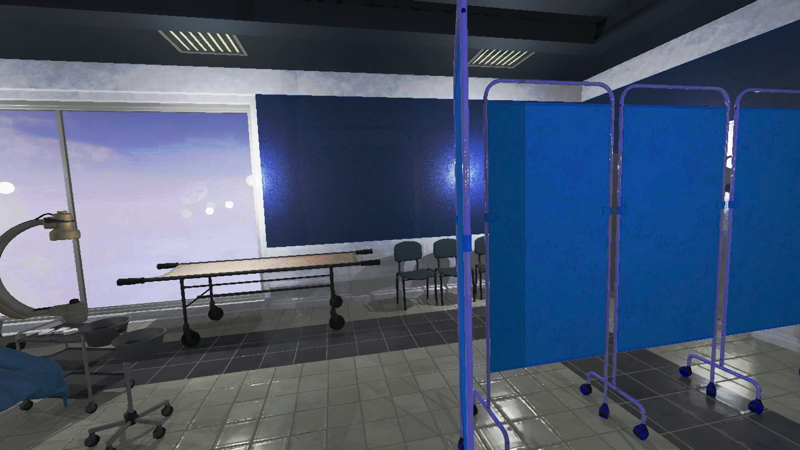} & \includegraphics[width=\linewidth, frame]{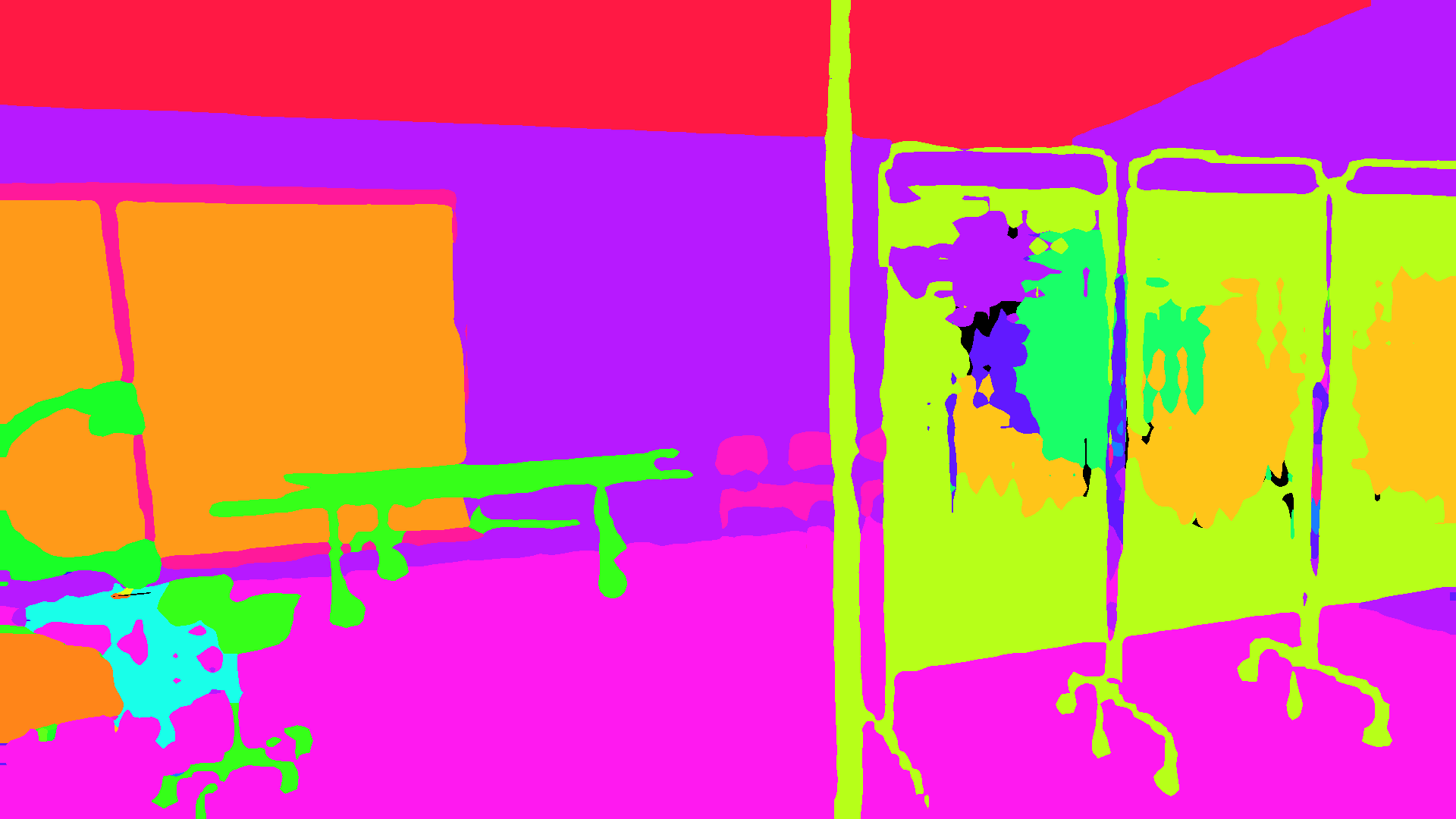} & \includegraphics[width=\linewidth, frame]{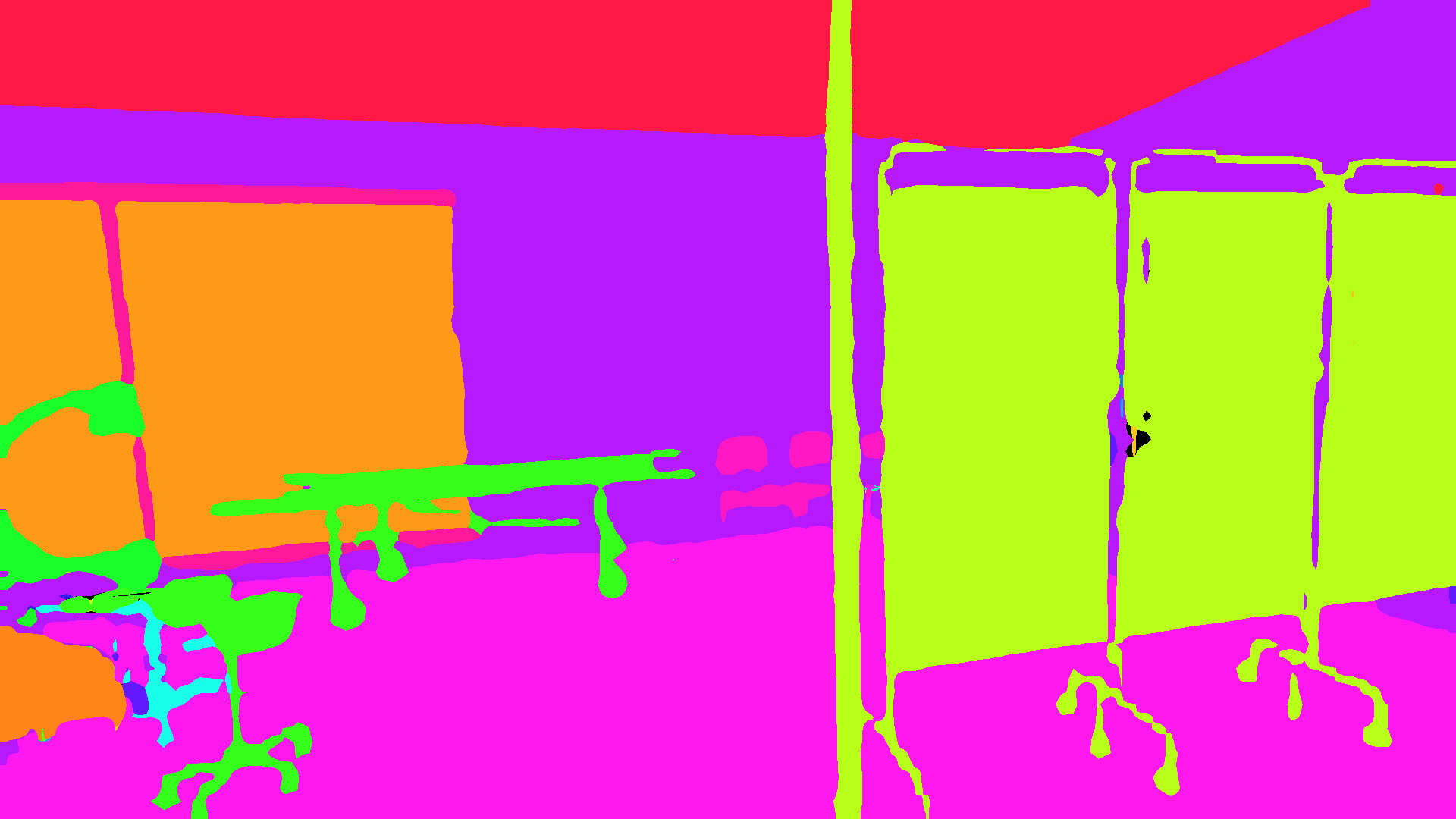} & \includegraphics[width=\linewidth, frame]{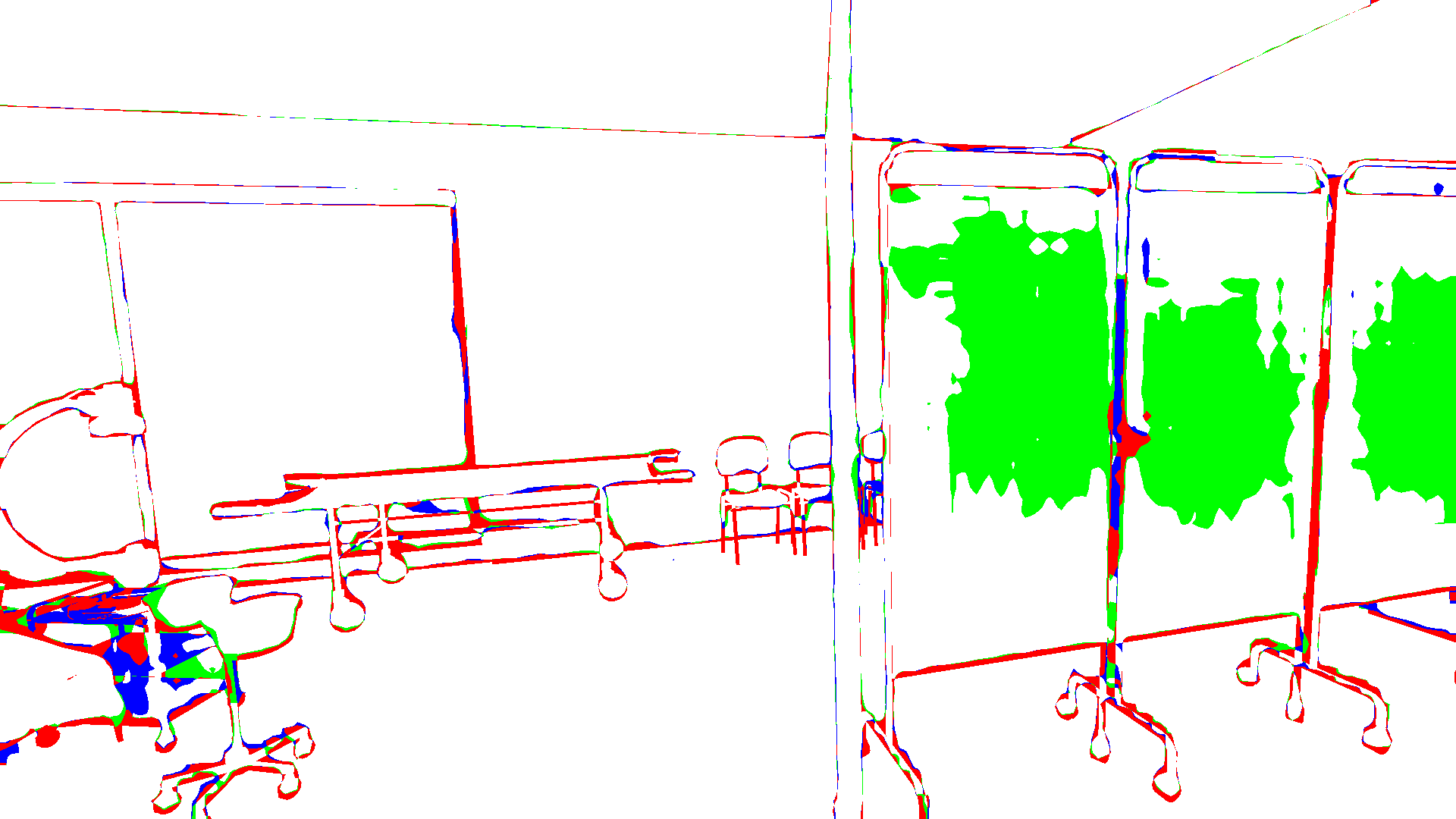}\\
\\
\end{tabular}
}
\caption{Semantic segmentation results on Syn-Mediverse 55-44 (3 tasks).}
\label{fig:qual-analysisSS}
\end{subfigure}
\caption{Qualitative results of \net~in comparison to the best-performing baselines for CISS on the Endovis18 and Syn-Mediverse dataset. The improvement/error maps show pixels misclassified by the baseline and correctly predicted by \net~in green and vice-versa in blue. Incorrect predictions of both models are colored in red.}
\end{figure*}

We present qualitative evaluations of \net~with the best-performing baseline on Endovis18 and Syn-Mediverse in~\figref{fig:qual-analysisIS} and \figref{fig:qual-analysisSS}. We observe that both TOPICS and \net~yield accurate segmentation on jaws while \net~produces more complete segmentation across various tissue types \figref{fig:qual-analysisIS}(i, iii) and instrument shafts \figref{fig:qual-analysisIS}(ii). On Syn-Mediverse, we observe that \net~better differentiates fine structures of a wheelchair in \figref{fig:qual-analysisSS}(ii) which we attribute to our improved hierarchical dice loss. Further, it also achieves improved consistent prediction on large segments on person and medical curtain in \figref{fig:qual-analysisSS}(i) and \figref{fig:qual-analysisSS}(iii).


\subsection{Ablation Study on Influence of Different Components}

\begin{wraptable}{r}{0.5\textwidth}
\vspace{-1cm}
\centering
\scriptsize
\setlength{\tabcolsep}{4pt}
\caption{Ablation study on \net components. Results on MM-OR in mIoU (\%). $PL_H$: hierarchical pseudo-labeling; Curv: curvature ($c$), $\mathcal{L}_{D}$: hierarchical dice loss.}
\label{tab:ab_elem}
\begin{tabular}{c|cc|ccc}
 \toprule
 \multicolumn{6}{c}{\textbf{MM-OR 11-4-3-2-3}}\\
\textbf{Curv} & $\mathcal{L}_{D}$ & $PL_H$ & 1–14 & 15–19 & all \\
\midrule
2 & & & \textbf{50.38} & 31.88 & 40.12\\
\midrule
\multirow{3}{*}{3} & & & 50.16 & 33.87 & 41.96 \\
& \checkmark & & 44.13 & 36.22 & 40.89\\
& \checkmark & \checkmark & 50.20 & \textbf{38.52} & \textbf{45.10}\\
\bottomrule
\end{tabular}
\vspace{-.7cm}
\end{wraptable}

We show the influence of different components for the refinement CISS setting on the MM-OR dataset in \tabref{tab:ab_elem}. 

A higher curvature contributes an $1.99$pp increase on novel classes as it increases relative distances in hyperbolic space, allowing novel class prototypes to be positioned farther from existing ones with fewer updates and thus improving hierarchical separation.
This is further boosted by $2.35$pp with our hierarchical dice loss designed to mitigate class imbalance.
Also, our hierarchical pseudo-labeling elevates the base class performance by $6.07$pp when combined with our hierarchical dice loss, labeling ancestor classes and reducing false-negative background predictions.

\section{Conclusion}\label{sec:conclusion}
We present \net, a replay-free Class-Incremental Semantic Segmentation approach for continual robot-assisted surgical scene segmentation. We incorporate the dice loss for hierarchical hyperbolic segmentation with class imbalances and introduce hierarchical pseudo-labeling to improve accuracy across diverse backgrounds. Our approach outperforms state-of-the-art CISS methods on six continual robotic surgery environment settings. This underscores the value of hierarchical modeling in continual learning.

{\parskip=3pt\noindent\textbf{Prospect of Application}: 
Our approach advances ongoing efforts in developing automatic segmentation methods for surgical robotics. It allows robots to incrementally learn to segment new tools and tissues encountered during complex operations without the need for storing prior data. This continuous learning capability is crucial for developing autonomous surgical assistants that can seamlessly integrate into privacy-concerned dynamic surgical environments.}

\subsubsection{\ackname}
This work was partly funded by the Deutsche Forschungsgemeinschaft (DFG, German Research Foundation) – SFB 1597 – 499552394. Ema Mekic is supported by the Konrad Zuse School of Excellence in Learning and Intelligent Systems (ELIZA) through the DAAD programme Konrad Zuse Schools of Excellence in Artificial Intelligence, sponsored by the Federal Ministry of Education and Research.
\subsubsection{\discintname}
The authors have no competing interests to declare that are relevant to the content of this article.

%
%

{\footnotesize
\bibliographystyle{splncs04}
\bibliography{references}
}
\end{document}